\documentclass[journal]{IEEEtran}

\usepackage[margin=0.75in]{geometry}
\usepackage{graphicx}
\usepackage{booktabs}
\usepackage[table]{xcolor}
\usepackage{multirow}
\usepackage{amsmath,amssymb}
\usepackage{caption}
\usepackage{subcaption}
\usepackage[normalem]{ulem}
\usepackage{array}

\definecolor{accentcyan}{RGB}{220,242,245}
\definecolor{absrel}{RGB}{255,255,255}
\definecolor{delta}{RGB}{255,255,255}
\definecolor{ours}{RGB}{220,242,245}
\definecolor{framesetting}{RGB}{238,238,238}
\definecolor{alignsetting}{RGB}{248,248,248}
\definecolor{posemetric}{RGB}{255,255,255}

\usepackage{booktabs}
\usepackage{multirow}
\usepackage[table]{xcolor}
\usepackage[normalem]{ulem}
\usepackage{cite}
\usepackage[hidelinks]{hyperref}
\usepackage{amsmath}
\usepackage{amssymb}
\usepackage{bm}
\usepackage{algorithm}
\usepackage{algorithmic}
\usepackage[table]{xcolor}
\usepackage{caption}
\definecolor{accentcyan}{RGB}{225,245,248}

\ifCLASSINFOpdf
\else
\fi
\begin{document}
\bstctlcite{BSTcontrol}
%
\title{BlazeVGGT: 13× Faster VGGT at 2,000 Frames via Camera Token without Training}
%
%
%

\author{
Shujun~Yao$^{*}$,
Junhui~Lin$^{*}$,
Congying~Han,
Tiande~Guo,
Ziwen~Liu$^{\dagger}$
and~Bonan~Li$^{\dagger}$

\thanks{Shujun Yao, Congying Han, Tiande Guo and Bonan Li are with University of Chinese Academy of Sciences,
Beijing, China.}
\thanks{Junhui Lin is with the University of Maryland, College Park, MD, USA.}
\thanks{Ziwen Liu is with Beijing University of Posts and Telecommunications, Beijing, China.}
\thanks{$^{*}$ Equal contribution}
\thanks{$^{\dagger}$Corresponding author: Ziwen Liu (liuziwen@bupt.edu.cn) and Bonan Li (libonan@ucas.ac.cn)}
}

%
%

\markboth{Under Review}
{Shell \MakeLowercase{\textit{et al.}}: BlazeVGGT: Camera-Token DPP for Efficient Long-Sequence Visual Geometry Inference}
%



\IEEEaftertitletext{%
\begin{center}
\refstepcounter{figure}
\label{fig:teaser}
\vspace{-1cm}

\includegraphics[width=1\textwidth]{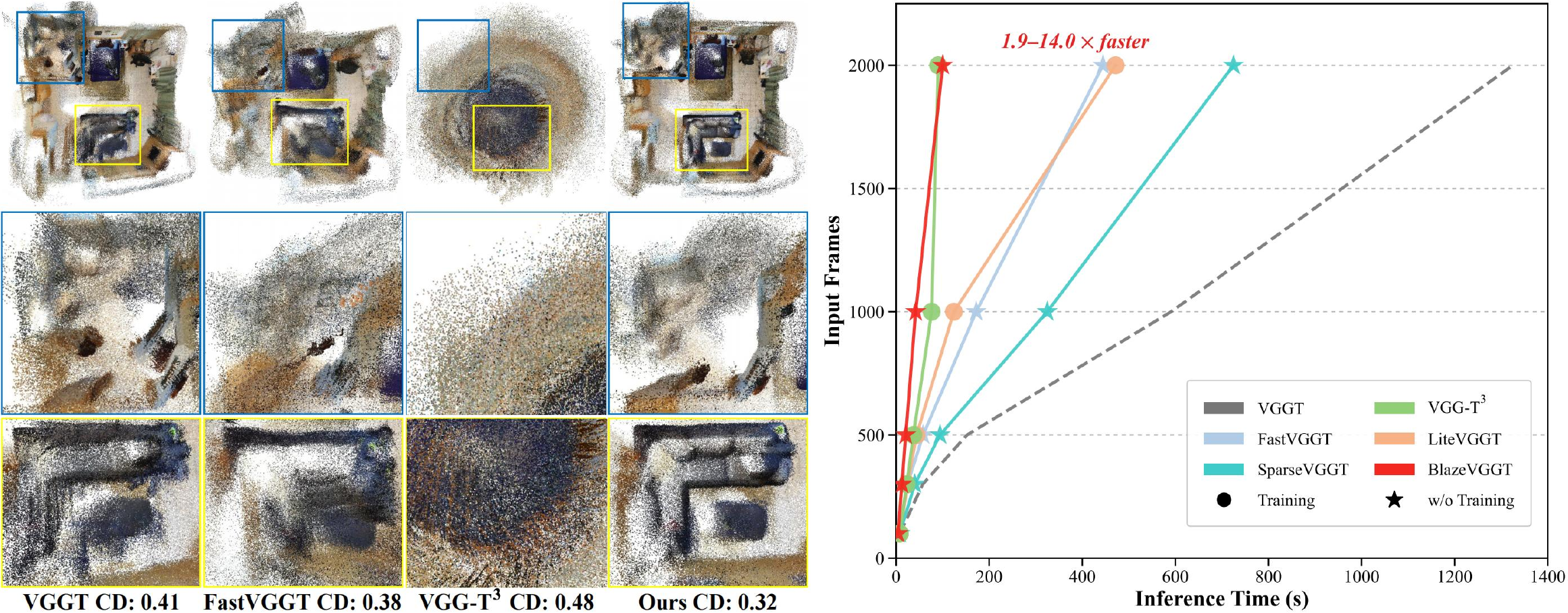}
\end{center}

\vspace{-0.5em}

\noindent
\small
Fig.~\thefigure.
We propose \textbf{BlazeVGGT}, a training-free frame-level KV sparsification framework that processes 2,000 images in a single inference, achieving $13.0\times$ faster inference while producing cleaner geometric reconstructions. All experiments use the VRAM-efficient implementation of VGGT~\cite{wang2025vggt} as the base model, following prior state-of-the-art methods for long-sequence evaluation.
}
\maketitle


\begin{abstract}

Feed-forward visual geometry models, exemplified by VGGT, have demonstrated strong multi-view reconstruction capabilities through dense global attention. However, the resulting all-to-all cross-view computation scales quadratically with the number of input views, making long-sequence inference particularly expensive due to substantial frame-level redundancy among highly overlapping observations.
We uncover a topology-first, geometry-later evolution in VGGT camera tokens: shallow-layer tokens reliably capture local view neighborhoods, whereas deeper-layer tokens progressively encode globally organized inter-view geometry.
Motivated by this observation, we introduce \textbf{BlazeVGGT}, a training-free framework for structure-aware frame-level key--value sparsification.
At each global-attention block, BlazeVGGT constructs a view-level similarity kernel from camera-token relations and employs greedy determinantal point process inference to select a compact set of mutually complementary views.
The selected views constitute the key--value context, while all input frames are retained as queries.
This asymmetric attention design preserves dense per-frame predictions while eliminating redundant cross-view computation, allowing our method to be directly integrated into pretrained VGGT as a plug-and-play module without parameter updates.
Its advantages become increasingly pronounced with sequence length: at 2,000 frames, BlazeVGGT achieves approximately 13$\times$ faster inference than VGGT while reducing the Chamfer Distance from 0.41 to 0.32.
Extensive experiments demonstrate a favorable accuracy--efficiency trade-off in point-cloud reconstruction and camera-pose estimation, compatibility with existing VGGT accelerators such as FastVGGT and LiteVGGT, and transferability to VGGT-based models including StreamVGGT and PAGE-4D.Codes are publicly available at https://github.com/MagiciannnY/BlazeVGGT.git.

\end{abstract}

\begin{IEEEkeywords}
Visual geometry transformer, Long-sequence reconstruction, Determinantal point process.
\end{IEEEkeywords}

%
\IEEEpeerreviewmaketitle

\vspace{-0.4cm}
\section{Introduction}
\label{sec:introduction}

\begin{figure}[t]
\centering
\includegraphics[width=\columnwidth]{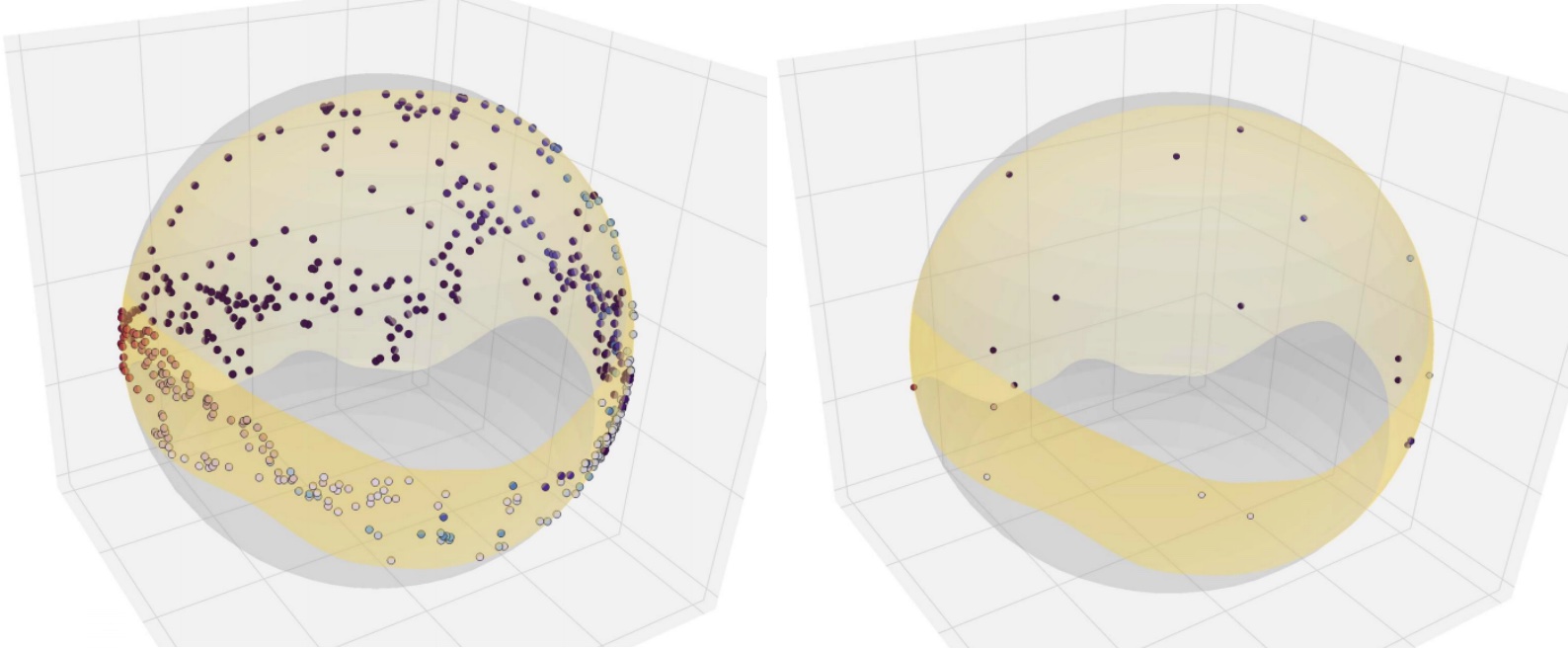}\\[-0.5ex]
\textbf{(a)}\\[2ex]
\includegraphics[width=\columnwidth]{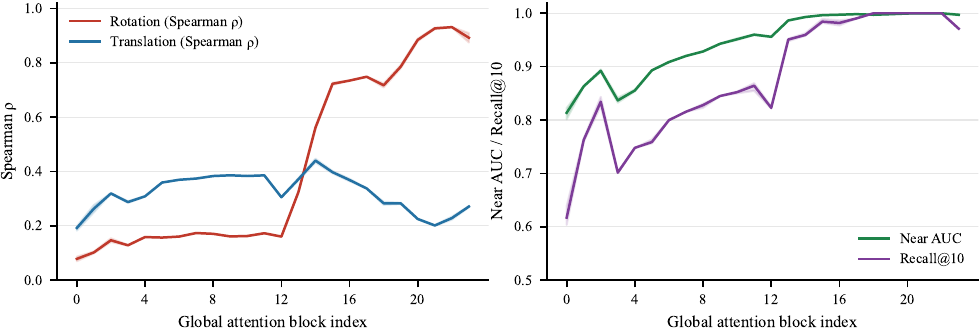}\\[-0.5ex]
\textbf{(b)}
\caption{Camera-token structure and its relation to camera geometry. (a)~Camera tokens from deep layer, projected to three dimensions by PCA and colored by camera yaw. The left view contains all frames; the right view contains the DPP-selected frames~\cite{Kulesza_2012}, showing that the selected views cover the yaw-organized token manifold. (b)~Layerwise diagnostics on a ScanNet~\cite{dai2017scannet} sequence. Left: Spearman correlation between centered-cosine camera-token distance and ground-truth rotation or translation distance. Right: local topology preservation measured by Near AUC and Recall@10. Shaded bands span the three token variants (concat, frame, global).}
\label{fig:motivation_rsa}
\vspace{-0.6cm}
\end{figure}

Feed-forward visual geometry models have recently emerged as powerful alternatives to conventional multi-stage reconstruction pipelines.
In particular, VGGT~\cite{wang2025vggt} jointly estimates camera parameters, depth maps, point maps, and point tracks from an arbitrary collection of input views within a single feed-forward network.
Its strong multi-view reasoning capability is largely enabled by global attention, which facilitates information exchange across all input views.
However, the resulting all-to-all cross-view computation scales unfavorably with the number of input frames.
Given $S$ views and $P$ tokens per view, each global-attention block operates over $SP$ tokens and incurs a dominant computational complexity of $\mathcal{O}(S^2P^2C)$, where $C$ denotes the feature dimension.
This quadratic dependence on $S$ becomes prohibitively expensive for long image sequences, which commonly contain substantial redundancy among highly overlapping observations.

Existing VGGT acceleration methods~\cite{shen2025fastvggt,litevggt2025,come2025,vggttt2026,httm2025,sparsevggt2025,flashvggt2025} primarily reduce the global-attention workload by compressing token or KV representations, or sparsifying global attention.
Although effective, these approaches retain several limitations.
Some methods~\cite{come2025,litevggt2025,vggttt2026} require additional training, fine-tuning, or input-specific optimization, restricting their direct applicability to pretrained models.
Training-free alternatives~\cite{shen2025fastvggt,httm2025,sparsevggt2025,flashvggt2025} generally operate at the granularity of individual tokens, token descriptors, attention heads, or sparse attention blocks, without explicitly modeling the structured redundancy among complete views.
Consequently, multiple near-duplicate frames may continue to serve as cross-view information sources even after their spatial representations have been compressed.
Moreover, aggressive token-level reduction may fragment the coherent intra-frame evidence required for geometric reasoning.
These limitations motivate a complementary perspective: rather than determining which individual tokens should be retained, we ask which views are necessary to provide a representative and geometrically complementary global context.

This view-level perspective is particularly relevant to multi-view geometry, where redundancy exhibits a strong frame-wise structure.
Adjacent views frequently capture nearly identical scene content, whereas a small number of geometrically distinct viewpoints may provide substantially more complementary evidence.
Naive frame sampling, however, can overlook geometrically important observations and fail to represent the global scene structure.
An effective view-selection mechanism must therefore preserve each selected frame as a complete unit while identifying a compact subset that jointly covers the input view distribution.
As illustrated in Fig.~\ref{fig:motivation_rsa}(a), the input views densely populate the camera-token space with considerable redundancy, whereas a compact subset can span its underlying view structure.
The key challenge is to identify such representative views during inference, particularly in shallow layers where reliable camera geometry has not yet been recovered.

To determine whether the intermediate representations of VGGT provide an effective signal for view selection, we systematically analyze the relational structure encoded by camera tokens at every global-attention block and compare it with ground-truth camera geometry.
As shown in Fig.~\ref{fig:motivation_rsa}(b), camera tokens exhibit a characteristic topology-first, geometry-later evolution.
Even in shallow layers, they achieve high Near AUC and Recall@10, indicating that local view neighborhoods are already reliably preserved.
At the same time, their rank correlation with global camera geometry remains limited: shallow camera tokens can distinguish nearby views from non-neighboring ones but do not yet accurately encode their global arrangement.
Around the middle of the network, this behavior changes markedly.
The camera-token similarity structure becomes increasingly consistent with global view geometry, particularly with respect to camera rotation, whereas metric translation remains less precisely calibrated.
These observations suggest that camera tokens progressively evolve from neighborhood-aware frame descriptors into task-relevant geometric representations, rather than explicitly recovering complete metric 6-DoF camera poses.

This progressive representation is particularly well aligned with the requirements of frame-level attention sparsification.
In shallow layers, accurate global geometry is unnecessary for redundancy suppression: identifying that multiple views belong to the same local neighborhood is already sufficient to avoid repeatedly attending to nearly duplicate observations.
In deeper layers, where cross-view reasoning benefits from a more globally representative context, camera tokens have developed a substantially more organized inter-view geometry.
We therefore arrive at a simple principle: local topology is sufficient for redundancy suppression, whereas global geometry promotes scene-wide view coverage.
Importantly, both forms of structure emerge directly from the camera tokens of the pretrained network, eliminating the need for an external pose estimator or an additional feature encoder.

Motivated by these observations, we introduce BlazeVGGT, a training-free framework for structure-aware frame-level key--value (KV) sparsification.
At each global-attention block, BlazeVGGT constructs a view-level similarity kernel from the evolving relations among camera tokens and employs greedy determinantal point process (DPP) selection~\cite{Kulesza_2012} to identify a compact set of mutually complementary views.
Unlike independent view ranking, the DPP criterion jointly accounts for redundancy and diversity within the selected set.
As camera-token representations evolve across layers, the same selection mechanism naturally transitions from suppressing local view redundancy in shallow layers to promoting broader scene coverage in deeper layers. The selected views are used exclusively to construct the KV context, while all input frames are retained in the query path.
Once a view is selected, its complete token block is preserved, thereby maintaining the coherent intra-frame spatial evidence required for geometric reasoning.
This query--KV asymmetry preserves predictions for every input frame while substantially reducing redundant cross-view computation.
Specifically, if $K_\ell \ll S$ views are selected at the $\ell$-th global-attention block, the dominant attention complexity is reduced from $\mathcal{O}(S^2P^2C)$ to $\mathcal{O}(SK_\ell P^2C)$.
Because BlazeVGGT modifies only the routing of the KV context and introduces no learnable parameters, it can be directly integrated into pretrained VGGT models as a plug-and-play module without retraining.
Its frame-level operation is also complementary to existing acceleration techniques that reduce computation at the token or attention level.

Extensive experiments demonstrate that BlazeVGGT achieves a favorable accuracy--efficiency trade-off, with increasingly pronounced gains as the sequence length grows.
As shown in Fig.~\ref{fig:teaser}, at 2,000 input frames, our method reduces the inference time from more than 1,300s to approximately 100s, corresponding to a speedup of about $13\times$, while reducing the Chamfer Distance from 0.41 to 0.32.
BlazeVGGT also maintains competitive performance in point-cloud reconstruction and camera-pose estimation.
Moreover, it can be seamlessly combined with existing VGGT accelerators, including FastVGGT~\cite{shen2025fastvggt} and LiteVGGT~\cite{litevggt2025}, and transferred to other VGGT-based architectures and inference settings, including StreamVGGT~\cite{zhuo2025streaming} and PAGE-4D~\cite{ICLR2026_3d551707}.
Collectively, these results demonstrate the scalability, compatibility, and generalizability of the proposed approach.

Our contributions are summarized as follows:
\begin{itemize}
    \item We uncover a topology-to-geometry evolution in VGGT camera tokens: local view neighborhoods emerge in shallow layers, while globally organized inter-view geometry strengthens in deeper layers.

    \item We introduce BlazeVGGT, a training-free framework for frame-level KV sparsification that combines camera-token relations with DPP-based view selection, retaining all queries and the complete token blocks of selected views.

    \item Experiments demonstrate favorable accuracy--efficiency trade-offs on sequences of up to 2,000 frames, compatibility with existing VGGT accelerators, and transferability across VGGT-based inference pipelines.
\end{itemize}

\vspace{-0.4cm}
\section{Related Work}

\subsection{3D Reconstruction}

Classical 3D reconstruction estimates scene structure and camera motion through feature correspondence, geometric estimation, and global optimization. Structure-from-motion (SfM) pipelines recover sparse geometry using matching, triangulation, and bundle adjustment~\cite{hartley_multiple_2000,snavely2006photo,frahm2010building,agarwal2011building,wu2013towards,schoenberger2016sfm,crandall2012sfm}. Their correspondence front ends have evolved from handcrafted descriptors and dense matching~\cite{mikolajczyk2005evaluation,liu2010sift,fan2019localfeatures} to learned descriptors~\cite{yi_lift_2016,detone2018self,dusmanu2019d2,tyszkiewicz2020disk}, learned matching~\cite{sarlin2020superglue,sun2021loftr,shen2023semidense,chen2022aspanformer,lindenberger2023lightglue,truong2023pdc}, and featuremetric refinement~\cite{sarlin2025pixsfm}, while classical solvers and registration remain fundamental~\cite{hartley1997defense,quan1999linear,gao2003complete,nister2004efficient,umeyama1991least,besl1992icp}. Dense multi-view stereo complements SfM through traditional patch-based methods~\cite{Furukawa2010MVS,shen2013patchbased,galliani2015massively,schonberger2016pixelwise,li2020cldmvs} and learned cost-volume formulations~\cite{yao2018mvsnet,gu2020cascade,li2023nrmvsnet,geomvsnet,xu2022multi}, although sparse wide-baseline and large-scale settings remain challenging~\cite{ji2021surfacenet,zhang2022gigamvs}. Neural rendering methods such as NeRF~\cite{mildenhall2021nerf} and 3D Gaussian Splatting~\cite{kerbl3Dgaussians} improve novel-view synthesis but typically require per-scene optimization. Direct and learned SfM systems~\cite{dso,zhou2017unsupervised,ummenhofer2017demon,tang2018ba,wei2020deepsfm,teed2018deepv2d,teed2021droid}, together with non-rigid formulations~\cite{torresani2008nonrigid}, further extend this paradigm, but generally retain iterative, pairwise, or sequential processing.

\vspace{-0.2cm}
\subsection{Feed-Forward Visual Geometry Models}

Feed-forward visual geometry models replace explicit per-scene optimization with learned geometric priors. DUSt3R~\cite{dust3r2024} predicts aligned point maps from image pairs, while MASt3R~\cite{mast3r2025} incorporates metric-aware matching. Fast3R~\cite{yang2025fast3r}, Spann3R~\cite{wang20243d}, and CUT3R~\cite{wang2025continuous} extend this paradigm to multi-view and persistent reconstruction. VGGT~\cite{wang2025vggt} instead jointly infers cameras, depth, point maps, and tracks by alternating frame-wise and global attention, enabling direct interaction across the entire image collection. StreamVGGT~\cite{zhuo2025streaming} adapts this architecture to continuous video, while VGGT-Long~\cite{vggtlong2025}, IncVGGT~\cite{incvggt2026}, and MERG3R~\cite{merg3r2026} address long-range reconstruction through chunking, incremental processing, or divide-and-conquer. Nevertheless, benchmarks such as KITTI-360~\cite{liao2022kitti} highlight the continuing need for scalable global reasoning over large image collections.
\vspace{-0.2cm}
\subsection{Efficient VGGT}

Efficient VGGT methods can be broadly divided according to whether they require adaptation of the pretrained model. Optimization-dependent approaches include VGG-T$^3$~\cite{vggttt2026}, which compresses variable-length KV representations through test-time training; LiteVGGT~\cite{litevggt2025}, which fine-tunes the model for geometry-aware token merging; and Co-Me~\cite{come2025}, which trains an auxiliary merging controller. Although differing in where adaptation occurs, these methods require either offline training or input-specific optimization. Training-free methods instead modify inference while preserving the pretrained parameters. FastVGGT~\cite{shen2025fastvggt} and HTTM~\cite{httm2025} reduce redundancy through global or head-wise token merging, whereas SparseVGGT~\cite{sparsevggt2025} adopts adaptive block-sparse attention. Related approaches use descriptor compression, sensitivity-aware sparsity allocation, or redesigned global attention~\cite{flashvggt2025,hess2026,avggt2025}. These methods primarily operate on spatial tokens, attention heads, or connectivity patterns. BlazeVGGT is likewise training-free but targets redundancy among complete views, restricting the KV context to geometrically complementary frames while retaining all input queries and the full token block of each selected view. Its view-level operation is therefore complementary to token merging, sparse attention, and model adaptation.

\vspace{-0.4cm}
\section{Methodology}
\label{sec:method}

\begin{figure*}[t]
\centering
\includegraphics[width=\textwidth]{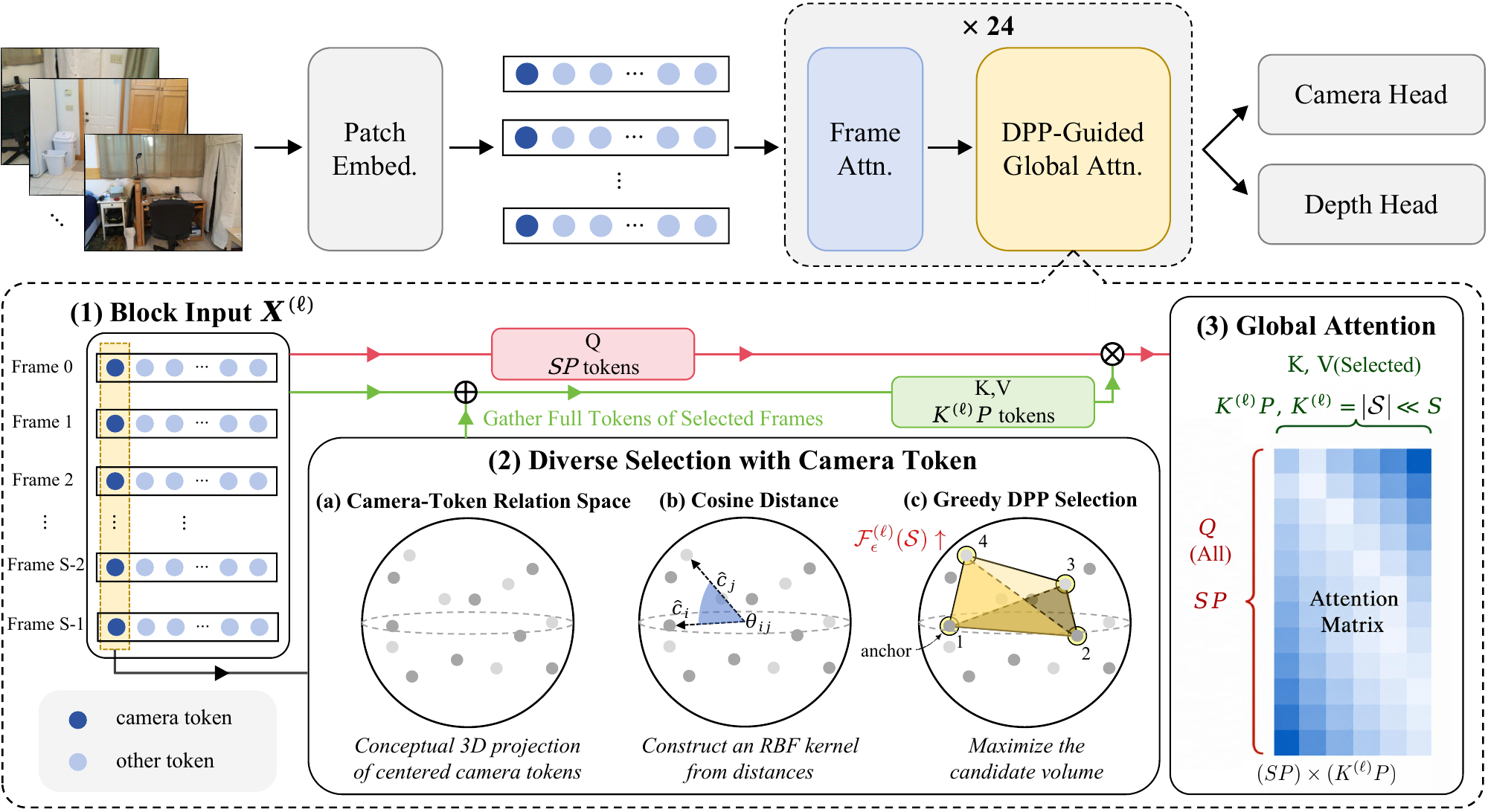}
\caption{Overview of BlazeVGGT. At each global-attention block, camera tokens from all input frames define a view-level similarity structure, from which greedy DPP~\cite{Kulesza_2012} selection retains a diverse subset of frames. All frames remain in the Query path, while the selected frame blocks provide the Key--Value context for global attention; the resulting features are passed to the camera and depth heads.}
\label{fig:method}
\vspace{-0.55cm}
\end{figure*}

\subsection{Preliminary}
\label{sec:preliminary}

We build upon Visual Geometry Grounded Transformer (VGGT)~\cite{wang2025vggt},
a feed-forward model that jointly predicts camera parameters, depth maps, point
maps, and point tracks from a set of RGB images.
Let $\Omega=\{0,\ldots,S-1\}$ denote the index set of $S$ input views
$\{\bm I_i\}_{i\in\Omega}$, where $\bm I_i\in \mathbb{R}^{H\times W\times 3}$.
VGGT first independently encodes
each image with DINOv2~\cite{oquab2024dinov} into a sequence of patch tokens.
Each view is further augmented with one learnable camera token and four register
tokens.

At the input to the $\ell$-th global-attention block, we denote the token block
associated with view $i$ as
\begin{equation}
    \bm X_i^{(\ell)}
    =
    \left[
        \bm c_i^{(\ell)};
        \bm R_i^{(\ell)};
        \bm Z_i^{(\ell)}
    \right]
    \in \mathbb{R}^{P \times C},
    \label{eq:view_tokens}
\end{equation}
where $\bm c_i^{(\ell)} \in \mathbb{R}^{C}$ is the camera token,
$\bm R_i^{(\ell)} \in \mathbb{R}^{4 \times C}$ denotes the register tokens,
$\bm Z_i^{(\ell)} \in \mathbb{R}^{(P-5) \times C}$ denotes the patch tokens,
$P$ is the total number of tokens per view, and $C$ is the feature dimension.
The camera tokens are progressively updated throughout the backbone and are
ultimately decoded by the camera head to predict the corresponding camera
parameters.

\noindent\textbf{\textit{Alternating frame and global attention.}}
The VGGT aggregator alternates between frame-wise and global self-attention.
Frame-wise attention processes each view independently, whereas global attention
jointly processes tokens from all views to enable cross-view information
exchange.
For the $\ell$-th global-attention block, the per-view token blocks are
concatenated as
\begin{equation}
    \bm X^{(\ell)}
    =
    \left[
        \bm X_0^{(\ell)};
        \ldots;
        \bm X_{S-1}^{(\ell)}
    \right]
    \in \mathbb{R}^{SP\times C}.
\end{equation}
After linear projection, the query, key, and value representations are
partitioned accordingly by view, and standard global self-attention is computed as
\begin{equation}
    \operatorname{Attention}(\bm Q,\bm K,\bm V)
    =
    \operatorname{softmax}
    \left(
        \frac{\bm Q\bm K^{\top}}{\sqrt{d_h}}
    \right)\bm V,
    \label{eq:vggt_global_attn}
\end{equation}
where $d_h$ denotes the feature dimension of each attention head.
Thus, every query token attends to the key--value tokens from all $S$ views.
With a fixed number of tokens per view, the global-attention matrix contains
$(S\cdot P)^2$ pairwise interactions, and its computational cost therefore grows
quadratically with the number of input views.
This dense cross-view interaction is the primary target of our methods introduced below.

\vspace{-0.2cm}
\subsection{Our Method}
BlazeVGGT exploits the progressively emerging inter-view structure of VGGT camera tokens to reduce redundant cross-view computation.
As illustrated in Fig.~\ref{fig:method}, our framework consists of two simple steps.
First, we construct a view-level similarity kernel from camera tokens and select a compact set of mutually complementary views using a determinantal objective.
Second, we retain all input frames as Queries while restricting the KV context of global attention to the selected views.
The entire procedure is training-free and operates directly on the intermediate representations of a pretrained VGGT.

\noindent\textbf{\textit{Camera-Token Guided View Selection.}}
Our analysis in Sec.~\ref{sec:introduction} shows that pairwise relations among camera tokens provide a progressively richer representation of inter-view structure.
Rather than decoding explicit camera poses, BlazeVGGT directly uses this relational structure to determine which views provide redundant or complementary evidence.

To emphasize relative differences among views while suppressing the component shared across the sequence, we first center the camera tokens across the input sequence and normalize them,
\begin{equation}
    \hat{\bm c}_i^{(\ell)}
    =
    \frac{
        \bm c_i^{(\ell)}
        -
        \bar{\bm c}^{(\ell)}
    }{
        \left\|
        \bm c_i^{(\ell)}
        -
        \bar{\bm c}^{(\ell)}
        \right\|_2
    }
    \label{eq:camera_normalization}
\end{equation}
with $\bar{\bm c}^{(\ell)} = \frac{1}{S}\sum_{j=0}^{S-1}\bm c_j^{(\ell)}$.
The dissimilarity between two views is measured by their cosine distance,
\begin{equation}
    d_{ij}^{(\ell)}
    =
    1-
    \left(
        \hat{\bm c}_i^{(\ell)}
    \right)^\top
    \hat{\bm c}_j^{(\ell)}.
    \label{eq:camera_distance}
\end{equation}
We convert these distances into an inter-view similarity kernel
\begin{equation}
    L_{ij}^{(\ell)}
    =
    \exp
    \left(
        -\frac{d_{ij}^{(\ell)}}{s^{(\ell)}}
    \right),
    \label{eq:view_kernel}
\end{equation}
where $s^{(\ell)}$ is an adaptive scale estimated from the pairwise distances at the current block.
Since the normalized camera-token distance is equivalent to squared Euclidean distance on the unit sphere, Eq.~\eqref{eq:view_kernel} defines a positive-semidefinite RBF kernel.

\begin{algorithm}[t]
\caption{Camera-Token View Selection in BlazeVGGT}
\label{alg:view_selection}
\begin{algorithmic}[1]
\REQUIRE Camera tokens from $S$ views at the current block, threshold $\epsilon$, maximum view budget $K$
\ENSURE Selected view-index set $\mathcal{S}$ for the KV context

\STATE Center and normalize camera tokens.
\STATE Build the adaptive RBF kernel $\bm L$ (Eq.~\eqref{eq:view_kernel}).
\STATE $\mathcal{S} \leftarrow \{0\}$ (Keep the anchor view)
\STATE Initialize the Cholesky factor of $\bm L_{\mathcal{S}}$.
\STATE Compute residual scores $r_j$ using Eq.~\eqref{eq:marginal_gain}.
\WHILE{$|\mathcal{S}| < \min(K,S)$}
    \STATE Select the unselected view $j$ with the largest $r_j$.
    \IF{$r_j < \epsilon$}
        \STATE \textbf{break}
    \ENDIF
    \STATE $\mathcal{S} \leftarrow \mathcal{S} \cup \{j\}$
    \STATE Apply incremental Cholesky updates (Eqs.~\eqref{eq:chol_coefficient}--\eqref{eq:residual_update}).
\ENDWHILE
\RETURN $\mathcal{S}$
\end{algorithmic}
\end{algorithm}

Given the inter-view kernel, the remaining question is how to select a
compact set of views that are collectively informative. Independently
ranking views by an importance score can retain multiple observations
that are individually informative but mutually redundant. We instead
model view selection as a set-level diversity problem. Following the diversity principle of determinantal point processes
(DPPs) \cite{Kulesza_2012}, we define the diversity of a
view subset $\mathcal{S}\subseteq\Omega$ at block $\ell$ as
$\mathcal{D}^{(\ell)}(\mathcal{S})
:= \det(\bm L_{\mathcal{S}}^{(\ell)})$.
Geometrically, the determinant measures the
volume spanned by the selected views in the kernel feature space.
Highly correlated views contribute little additional volume, whereas
views that are dissimilar to the current set provide larger
complementary coverage.

Because $L_{ii}^{(\ell)}=1$, Hadmard's inequality give
$\mathcal{D}^{(\ell)}(\mathcal{S})\leq 1$, whereas every singleton
has determinant one. Unconstrained determinant maximization therefore admits a
singleton solution and cannot by itself determine the adaptive subset size. We
instead interpret the implemented stopping rule through the threshold-calibrated
log-determinant objective, with $0<\epsilon\leq 1$,
\begin{equation}
\mathcal{F}_{\epsilon}^{(\ell)}(\mathcal{S})
=
\log\mathcal{D}^{(\ell)}(\mathcal{S})
-
\lvert\mathcal{S}\rvert\log\epsilon,
\label{eq:thresholded_logdet_objective}
\end{equation}
subject to $0\in\mathcal{S}$ and
$\lvert\mathcal{S}\rvert\leq K$.
We optimize this objective by greedy ascent. Consider adding a candidate
view $j\in\Omega\setminus\mathcal{S}$. By the Schur complement,

\begin{equation}
\mathcal{D}^{(\ell)}(\mathcal{S}\cup\{j\})
=
\mathcal{D}^{(\ell)}(\mathcal{S})\,r_j^{(\ell)},
\label{eq:det_gain}
\end{equation}

where
\begin{equation}
r_j^{(\ell)}
=
L_{jj}^{(\ell)}
-
\left(\bm l_j^{(\ell)}\right)^\top
\left(
\bm L_{\mathcal{S}}^{(\ell)}
\right)^{-1}
\bm l_j^{(\ell)},
\qquad
\bm l_j^{(\ell)}
=
\bm L_{\mathcal{S},j}^{(\ell)}.
\label{eq:marginal_gain}
\end{equation}

Thus, $r_j^{(\ell)}$ measures the marginal diversity contributed by
view $j$. Its marginal gain under Eq.~\eqref{eq:thresholded_logdet_objective} is
\begin{equation}
\mathcal{F}_{\epsilon}^{(\ell)}
\left(\mathcal{S}\cup\{j\}\right)
-
\mathcal{F}_{\epsilon}^{(\ell)}
\left(\mathcal{S}\right)
=
\log\frac{r_j^{(\ell)}}{\epsilon}.
\label{eq:thresholded_marginal_gain}
\end{equation}
Consequently, adding a view has nonnegative marginal gain exactly when
$r_j^{(\ell)}\geq\epsilon$. At each iteration, we select

\begin{equation}
j^\star
=
\arg\max_{j\in\Omega\setminus\mathcal{S}}
r_j^{(\ell)}.
\label{eq:greedy_selection}
\end{equation}

To avoid repeatedly evaluating the matrix inverse in
Eq.~\eqref{eq:marginal_gain}, we maintain the residuals using
incremental Cholesky updates\cite{chen2018fastgreedymapinference}.
Let
$\bm L_{\mathcal{S}}^{(\ell)}=\bm B\bm B^\top$
and
$\bm q_j=\bm B^{-1}\bm l_j^{(\ell)}$.
After selecting $j^\star$, the new Cholesky coefficient for a remaining
candidate $j$ is

\begin{equation}
e_j
=
\frac{
L_{j^\star j}^{(\ell)}
-
\bm q_{j^\star}^{\top}\bm q_j
}{
\sqrt{r_{j^\star}^{(\ell)}}
},
\label{eq:chol_coefficient}
\end{equation}
and its residual can be updated efficiently as

\begin{equation}
r_j^{(\ell)}
\leftarrow
r_j^{(\ell)}-e_j^2,
\qquad
j\in\Omega\setminus\mathcal{S}.
\label{eq:residual_update}
\end{equation}

Selection starts from the anchor view and greedily increases
$\mathcal{F}_{\epsilon}^{(\ell)}$ while the largest residual is at least
$\epsilon$, subject to the maximum budget $K$. This is a thresholded greedy
approximation and the resulting
cardinality is therefore adaptive: highly redundant
representations naturally retain fewer KV views, whereas more diverse
ones preserve a larger context. Algorithm~\ref{alg:view_selection}
summarizes the procedure.

Importantly, the same selection rule is used throughout the network.
In shallow blocks, where camera tokens primarily capture local
neighborhood structure, the determinant mainly suppresses redundant
nearby observations. As camera-token relations become increasingly
geometry-aware in deeper blocks, the same objective instead favors
views that provide complementary scene coverage. BlazeVGGT therefore
requires neither an explicit stage boundary nor separate shallow- and
deep-layer selectors; the change in selection behavior emerges directly
from the evolving camera-token representation.

\begin{table*}[t]
\centering
\caption{Point cloud reconstruction on ScanNet-50~\cite{dai2017scannet} with different numbers of input views. We report Chamfer Distance (CD) and inference time.
The 2,000-view setting uses two-GPU sequence parallelism, whereas the 1,000-view setting uses single-GPU inference. 
BlazeVGGT achieves 13.2$\times$ and 14.0$\times$ speedups in these two settings, respectively. 
The slightly lower speedup at 2,000 views may partly reflect inter-GPU communication and synchronization overhead, 
which was not separately profiled.}
\label{tab:scannet_recon}

\setlength{\tabcolsep}{10pt}
\renewcommand{\arraystretch}{1.25}

\resizebox{\textwidth}{!}{
\begin{tabular}{ll|cc|cc|cc|cc}
\specialrule{1pt}{0pt}{2pt}
\specialrule{0.3pt}{0pt}{0pt}
\multicolumn{2}{c|}{\multirow{2}{*}{Method}}
& \multicolumn{2}{c|}{2,000}
& \multicolumn{2}{c|}{1,000}
& \multicolumn{2}{c|}{500}
& \multicolumn{2}{c}{100} \\
\cline{3-10}
\multicolumn{2}{c|}{}
& CD$\downarrow$ & Time$\downarrow$
& CD$\downarrow$ & Time$\downarrow$
& CD$\downarrow$ & Time$\downarrow$
& CD$\downarrow$ & Time$\downarrow$ \\ 
\hline

\multicolumn{2}{c|}{VGGT\cite{wang2025vggt} (CVPR'25)}
& 0.4184 & 1325.1s
& 0.4760 & 591.0s
& 0.4646 & 152.3s
& 0.4377 & 8.1s \\ \hline

\multirow{2}{*}{Training}
& LiteVGGT\cite{litevggt2025} (CVPR'26)
& \textbf{0.3420} & 471.6s
& \textbf{0.4085} & 133.8s
& \textbf{0.3862} & 44.2s
& \textbf{0.3291} & 8.1s \\

& VGG-T$^3$\cite{vggttt2026} (CVPR'26)
& 0.4822 & \textbf{90.4s}
& 0.5142 & \textbf{76.3s}
& 0.4957 & \textbf{37.2s}
& 0.421 & \textbf{8.0s} \\
\hline

\multirow{3}{*}{w/o Training}
& FastVGGT\cite{shen2025fastvggt} (ICLR'26)
& \uline{0.3886} & \uline{444.5s}
& 0.4654 & \uline{172.7s}
& 0.4569 & \uline{58.0s}
& \uline{0.4234} & \uline{4.8s} \\

& SparseVGGT\cite{sparsevggt2025} (CVPR'26)
& {0.3923} & 725.0s
& \textbf{0.4247} & 324.8s
& \uline{0.4348} & 94.4s
& 0.4419 & 7.8s \\

\rowcolor{accentcyan}
\cellcolor{white} & BlazeVGGT
& \textbf{0.3282} & \textbf{100.1s} \textcolor{green!50!black}{(\scriptsize $\uparrow$13.2$\times$)}
& \uline{0.4283} & \textbf{42.1s} \textcolor{green!50!black}{(\scriptsize $\uparrow$14.0$\times$)}
& \textbf{0.4279} & \textbf{20.7s} \textcolor{green!50!black}{(\scriptsize $\uparrow$7.4$\times$)}
& \textbf{0.4111} & \textbf{4.2s} \textcolor{green!50!black}{(\scriptsize $\uparrow$1.9$\times$)} \\
\hline
\end{tabular}
}
\end{table*}

\begin{table*}[t]
\centering
\caption{
Point cloud reconstruction on 7 Scenes~\cite{shotton2013scene} and NRGBD~\cite{Azinovic_2022_CVPR} using stride-3
sampling. We report accuracy (Acc), completeness (Comp), normal
consistency (NC), and inference time.
}
\label{tab:recon_7scenes_nrgbd}

\scriptsize
\setlength{\tabcolsep}{7pt}
\renewcommand{\arraystretch}{1.25}

\resizebox{\textwidth}{!}{%
\begin{tabular}{ll|ccccccc|ccccccc}
\specialrule{1pt}{0pt}{2pt}
\specialrule{0.3pt}{0pt}{0pt}
\multicolumn{2}{c|}{\multirow{3}{*}{Method}}
& \multicolumn{7}{c|}{7 Scenes}
& \multicolumn{7}{c}{NRGBD} \\
\cline{3-9}
\cline{10-16}
&
& \multicolumn{2}{c}{Acc$\downarrow$}
& \multicolumn{2}{c}{Comp$\downarrow$}
& \multicolumn{2}{c}{NC$\uparrow$}
& \multirow{2}{*}{Time$\downarrow$}
& \multicolumn{2}{c}{Acc$\downarrow$}
& \multicolumn{2}{c}{Comp$\downarrow$}
& \multicolumn{2}{c}{NC$\uparrow$}
& \multirow{2}{*}{Time$\downarrow$} \\
\cline{3-8}
\cline{10-15}
&
& Mean & Med. & Mean & Med. & Mean & Med. &
& Mean & Med. & Mean & Med. & Mean & Med. & \\
\hline

\multicolumn{2}{c|}{VGGT\cite{wang2025vggt} (CVPR'25)}
& 0.0195 & 0.0183 & 0.0267 & 0.0199 & 0.5862 & 0.5829 & 65.9s
& {0.0287} & 0.0191 & {0.0177} & 0.0124 & 0.6646 & 0.6756 & 112.1s \\
\hline

\multirow{2}{*}{Training}
& LiteVGGT\cite{litevggt2025} (CVPR'26)
& \textbf{0.0203} & \textbf{0.0182} & \textbf{0.0249} & \textbf{0.0169} & \textbf{0.6065} & \textbf{0.6056} & \textbf{22.0s}
& \textbf{0.0354} & \textbf{0.0250} & \textbf{0.0255} & \textbf{0.0125} & \textbf{0.6908} & \textbf{0.6955} & \textbf{32.9s} \\

& VGG-T$^3$\cite{vggttt2026} (CVPR'26)
& 0.1306 & 0.0304 & 0.0701 & 0.0308 & 0.5858 & 0.5902 & 26.5s
& 0.1788 & 0.0525 & 0.1580 & 0.0171 & 0.6667 & 0.6713 & 34.2s \\
\hline

\multirow{3}{*}{w/o Training}
& FastVGGT\cite{shen2025fastvggt} (ICLR'26)
& \uline{0.0178} & \uline{0.0172} & \uline{0.0263} & \textbf{0.0193} & \uline{0.5895} & \uline{0.5907} & \uline{25.3s}
& 0.0315 & \uline{0.0180} & 0.0191 & 0.0126 & 0.6690 & \uline{0.6799} & \uline{47.2s} \\

& SparseVGGT\cite{sparsevggt2025} (CVPR'26)
& 0.0190 & 0.0187 & 0.0290 & 0.0207 & \textbf{0.6636} & \textbf{0.6666} & 45.6s
& \uline{0.0290} & 0.0259 & \uline{0.0185} & \uline{0.0117} & \textbf{0.8102} & \textbf{0.8439} & 71.5s \\

\rowcolor{accentcyan}
\cellcolor{white} & BlazeVGGT
& \textbf{0.0176} & \textbf{0.0170} & \textbf{0.0258} & \uline{0.0194} & 0.5859 & 0.5819 & \textbf{13.5s} \textcolor{green!50!black}{(\scriptsize $\uparrow$4.9$\times$)}
& \textbf{0.0163} & \textbf{0.0153} & \textbf{0.0169} & \textbf{0.0116} & \uline{0.6761} & 0.6694 & \textbf{14.8s} \textcolor{green!50!black}{(\scriptsize $\uparrow$7.6$\times$)} \\

\hline
\end{tabular}
}
\end{table*}

\noindent\textbf{\textit{Frame-Routed Global Attention.}} The selected views are used to sparsify only the source of cross-view information.
BlazeVGGT retains every input frame in the Query path, while the KV context is constructed only from the selected set
$\mathcal{S}^{(\ell)}$.
For brevity, the subscript $\mathcal{S}$ below refers to this layer-specific set.

For a single attention head, let
$\bm W_Q^{(\ell)},\bm W_K^{(\ell)},\bm W_V^{(\ell)}\in
\mathbb{R}^{C\times d_h}$ and
$K^{(\ell)}=\lvert\mathcal{S}^{(\ell)}\rvert$.
After flattening the frame and token dimensions,
$\bm X^{(\ell)}\in\mathbb{R}^{SP\times C}$ and
$\bm X_{\mathcal{S}}^{(\ell)}\in\mathbb{R}^{K^{(\ell)}P\times C}$.
We write
\begin{equation}
\begin{aligned}
    \bm Q^{(\ell)}
    &=
    \bm X^{(\ell)}
    \bm W_Q^{(\ell)},\\
    \bm K_{\mathcal{S}}^{(\ell)}
    &=
    \bm X_{\mathcal{S}}^{(\ell)}
    \bm W_K^{(\ell)},\\
    \bm V_{\mathcal{S}}^{(\ell)}
    &=
    \bm X_{\mathcal{S}}^{(\ell)}
    \bm W_V^{(\ell)}.
\end{aligned}
\label{eq:qkv_routing}
\end{equation}
Therefore,
$\bm Q^{(\ell)}\in\mathbb{R}^{SP\times d_h}$ and
$\bm K_{\mathcal{S}}^{(\ell)},
\bm V_{\mathcal{S}}^{(\ell)}
\in\mathbb{R}^{K^{(\ell)}P\times d_h}$.
The resulting global attention becomes
\begin{equation}
    \bm Y^{(\ell)}
    =
    \operatorname{softmax}
    \left(
        \frac{
            \bm Q^{(\ell)}
            \left(
                \bm K_{\mathcal{S}}^{(\ell)}
            \right)^\top
        }{
            \sqrt{d_h}
        }
    \right)
    \bm V_{\mathcal{S}}^{(\ell)}.
    \label{eq:routed_attention}
\end{equation}

This Q--KV asymmetry is central to BlazeVGGT.
Every input view remains a Query and therefore continues to receive cross-view updates and produce the original dense predictions.
Frame selection only determines which observations need to serve as \emph{cross-view evidence}. Moreover, sparsification is performed at the frame level rather than independently over spatial tokens.
Once a view is selected, all tokens associated with that view are preserved in the KV context.
BlazeVGGT therefore removes redundancy \emph{across views} without fragmenting the coherent spatial evidence \emph{within views}.

Full global attention involves $SP$ Queries and $SP$ KV tokens, giving a dominant attention complexity of
$\mathcal{O}(S^2P^2C)$. BlazeVGGT reduces the KV length to $K^{(\ell)}P$, yielding
\begin{equation}
    \mathcal{O}
    \left(
        S K^{(\ell)} P^2 C
    \right).
    \label{eq:complexity}
\end{equation}
When the retained-view budget remains bounded as the sequence grows, the dominant global-attention cost therefore scales approximately linearly rather than quadratically with the number of input views.

Since BlazeVGGT changes only the routing of the global KV context and introduces no learned parameters, it can be directly applied to pretrained VGGT models without retraining.
Its frame-level operation is also orthogonal to acceleration techniques that reduce redundancy at the token or attention level, making BlazeVGGT naturally composable with existing efficient VGGT variants.

\vspace{-0.4cm}
\section{Experiments}
\label{sec:experiments}

\subsection{Experimental Setup}
\label{sec:experimental_setup}

\noindent\textbf{\textit{Experimental Protocol.}}
Unless otherwise specified, VGGT in this section denotes the
VRAM-efficient implementation of VGGT~\cite{wang2025vggt} used in
our experiments, enabling larger inputs.
For the 2,000-frame comparison, following the long-sequence
implementation of VGG-T$^3$~\cite{vggttt2026}, we adapt the A800
evaluation implementations of VGGT, FastVGGT~\cite{shen2025fastvggt},
SparseVGGT~\cite{sparsevggt2025}, LiteVGGT~\cite{litevggt2025},
VGG-T$^3$, and BlazeVGGT to support inference on two NVIDIA
A800-SXM4-80GB GPUs connected by NVLink, while preserving their
respective algorithmic configurations. All remaining experiments are
conducted on a single NVIDIA A800 GPU, except that LiteVGGT-related
experiments outside the 2,000-frame comparison and high-frame-count
PAGE-4D experiments are conducted on an NVIDIA H20 GPU.

\noindent\textbf{\textit{Datasets and Evaluation Scope.}}
Our evaluation consists of main experiments and two complementary
generalization studies. The main experiments compare BlazeVGGT with
state-of-the-art methods for point cloud reconstruction on
ScanNet-50~\cite{dai2017scannet}, 7 Scenes~\cite{shotton2013scene},
and NRGBD~\cite{Azinovic_2022_CVPR}, and for camera-pose estimation on
ScanNet-50 under multiple input lengths. They further include ablations on
ScanNet-50 to isolate the effects of the selection threshold, the
Top-$K$ safety constraint, and the activated layer range.
Complementarily, we examine generalization in two directions: 
(1) seamless composition with existing VGGT accelerators on ScanNet-50
and Tanks \& Temples~\cite{knapitsch2017tanks}, and (2) direct transfer
to other VGGT-based models---StreamVGGT~\cite{zhuo2025streaming} for
online reconstruction on 7 Scenes and PAGE-4D~\cite{ICLR2026_3d551707}
for video-depth estimation on Sintel~\cite{Butler:ECCV:2012},
Bonn~\cite{palazzolo2019iros}, and DyCheck~\cite{gao2022dynamic}.

\begin{table*}[t]
\centering
\caption{
Camera pose estimation on ScanNet-50~\cite{dai2017scannet} under different input sequence
lengths. We report absolute trajectory error (ATE), absolute rotation
error (ARE), relative translation error (RPE-t), and relative rotation
error (RPE-r).
}
\label{tab:scannet_pose}

\scriptsize
\setlength{\tabcolsep}{4pt}
\renewcommand{\arraystretch}{1.10}

\resizebox{\textwidth}{!}{%
\begin{tabular}{ll|cccc|cccc|cccc|cccc}
\specialrule{1pt}{0pt}{2pt}
\specialrule{0.3pt}{0pt}{0pt}
\multicolumn{2}{c|}{\multirow{2}{*}{Method}}
& \multicolumn{4}{c|}{2,000}
& \multicolumn{4}{c|}{1,000}
& \multicolumn{4}{c|}{500}
& \multicolumn{4}{c}{100} \\
\cline{3-18}
\multicolumn{2}{c|}{}
& ATE$\downarrow$ & ARE$\downarrow$ & RPE-t$\downarrow$ & RPE-r$\downarrow$
& ATE$\downarrow$ & ARE$\downarrow$ & RPE-t$\downarrow$ & RPE-r$\downarrow$
& ATE$\downarrow$ & ARE$\downarrow$ & RPE-t$\downarrow$ & RPE-r$\downarrow$
& ATE$\downarrow$ & ARE$\downarrow$ & RPE-t$\downarrow$ & RPE-r$\downarrow$ \\
\hline

\multicolumn{2}{c|}{VGGT (CVPR'25)}
& 0.1644 & 7.0532 & 0.0346 & 0.9154
& 0.1336 & 5.6834 & 0.0419 & 1.1971
& 0.1236 & 4.8265 & 0.0414 & 0.9660
& 0.0921 & 3.9382 & 0.0562 & 1.2169 \\
\hline

\multirow{2}{*}{Training}
& LiteVGGT (CVPR'26)
& \textbf{0.1321} & \textbf{4.9931} & \textbf{0.0321} & \textbf{0.7468}
& \textbf{0.1497} & \textbf{7.4436} & \textbf{0.0387} & \textbf{1.1073}
& \textbf{0.1181} & \textbf{4.6390} & \textbf{0.0400} & \textbf{0.7811}
& \textbf{0.1077} & \textbf{4.1110} & \textbf{0.0732} & \textbf{1.4482} \\

& VGG-T$^3$ (CVPR'26)
& 0.9331 & 82.9325 & 0.1190 & 11.6955
& 0.5045 & 36.0913 & 0.1523 & 9.9699
& 0.5450 & 34.8302 & 0.1341 & 7.8155
& 0.2970 & 9.2451 & 0.1945 & 4.9533 \\
\hline

\multirow{3}{*}{w/o Training}
& FastVGGT (ICLR'26)
& \uline{0.2155} & \uline{10.1581} & \uline{0.0628} & {2.3282}
& \uline{0.1146} & \uline{4.0139} & \uline{0.0309} & \uline{0.6628}
& \textbf{0.0927} & \textbf{4.0140} & \textbf{0.0305} & \textbf{0.6418}
& \textbf{0.0932} & \textbf{3.8528} & \textbf{0.0600} & \textbf{1.2571} \\

& SparseVGGT (CVPR'26)
& 0.2712 & 14.2328 & 0.0649 & \uline{2.2359}
& 0.2205 & 9.2380 & 0.0566 & 1.8232
& 0.1270 & 5.0168 & 0.0467 & 1.1367
& 0.1047 & 4.4391 & \uline{0.0661} & 1.4633 \\

\rowcolor{accentcyan}
\cellcolor{white} & BlazeVGGT
& \textbf{0.1888} & \textbf{7.9601} & \textbf{0.0461} & \textbf{1.1116}
& \textbf{0.0910} & \textbf{3.8094} & \textbf{0.0270} & \textbf{0.5748}
& \uline{0.0954} & \uline{4.4483} & \uline{0.0322} & \uline{0.7136}
& \uline{0.1034} & \uline{4.1757} & 0.0675 & \uline{1.3618} \\

\hline
\end{tabular}
}
\vspace{-0.2cm}
\end{table*}

\begin{figure*}[t]
\centering
\includegraphics[width=\textwidth]{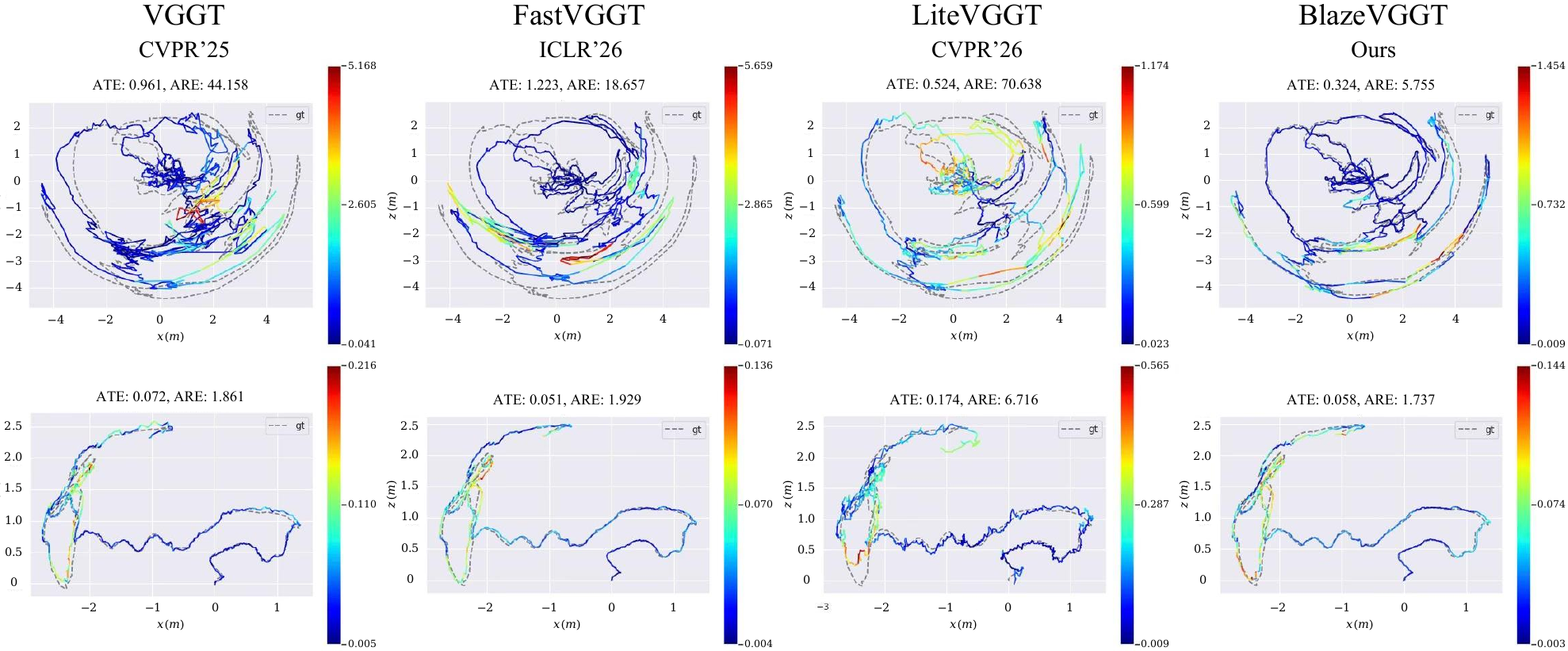}
\caption{Qualitative comparison of camera trajectory estimation in the
$x$--$z$ plane. The four columns correspond to VGGT~\cite{wang2025vggt}, FastVGGT~\cite{shen2025fastvggt}, LiteVGGT~\cite{litevggt2025},
and BlazeVGGT. }
\label{fig:camera_pose_visualization}
\vspace{-0.3cm}
\end{figure*}

\vspace{-0.2cm}
\subsection{Comparison with State-of-the-Art}
\label{sec:sota}
\noindent\textbf{\textit{Quantitative Comparison.}} We compare BlazeVGGT with VGGT~\cite{wang2025vggt} and four recent variants, including the training-dependent LiteVGGT~\cite{litevggt2025} and VGG-T$^3$~\cite{vggttt2026} and the training-free FastVGGT~\cite{shen2025fastvggt} and SparseVGGT~\cite{sparsevggt2025}. Evaluation covers point-cloud reconstruction and camera-pose estimation, with an emphasis on long-sequence inference.

\noindent{\textit{(\romannumeral 1) Point Cloud Reconstruction.}}
TABLES~\ref{tab:scannet_recon} and~\ref{tab:recon_7scenes_nrgbd}
demonstrate that inference efficiency is the most prominent advantage
of BlazeVGGT. On ScanNet-50, BlazeVGGT achieves the shortest inference
time among all compared methods at 100, 500, and 1,000 views, and
remains the fastest training-free method at 2,000 views. Moreover, its
speedup over VGGT grows rapidly as the input sequence becomes
longer, reaching 14.0$\times$ at 1,000 views and remaining above
13$\times$ under two-GPU inference at 2,000 views. Although the trained
VGG-T$^3$ is marginally faster in the 2,000-view setting, it produces
substantially less accurate reconstructions, highlighting the favorable
quality--efficiency trade-off achieved by BlazeVGGT without training or
input-specific optimization. This efficiency advantage is also
accompanied by competitive reconstruction quality: compared with the
trained variants, BlazeVGGT performs particularly well in the
longest-sequence setting, whereas LiteVGGT retains an advantage on
several shorter-sequence and surface-normal metrics. More directly,
among the training-free methods, BlazeVGGT achieves the best or
near-best distance-based reconstruction accuracy across the evaluated
sequence lengths while consistently requiring less inference time than
FastVGGT and SparseVGGT, with the efficiency gap becoming especially
pronounced for long sequences. Furthermore, the same behavior
generalizes beyond ScanNet-50: on both 7 Scenes and NRGBD, BlazeVGGT
records the lowest reported inference time while maintaining strong
geometric accuracy and completeness. SparseVGGT obtains higher normal
consistency in some settings; nevertheless, the overall results show
that BlazeVGGT substantially reduces the dominant cost of cross-view
attention without sacrificing competitive reconstruction quality.
Taken together, these results confirm that view-level KV sparsification
is particularly effective for long and highly redundant image
sequences, where it provides increasingly substantial computational
savings while preserving the geometric evidence required for dense
reconstruction.

\definecolor{ablred}{RGB}{255,220,220}
\definecolor{ablyellow}{RGB}{255,250,218}
\definecolor{ablgreen}{RGB}{220,250,225}

\definecolor{ablred}{RGB}{255,220,220}
\definecolor{ablyellow}{RGB}{255,250,218}
\definecolor{ablgreen}{RGB}{220,250,225}

\begin{table}[t]
\centering
\caption{
Ablation study of BlazeVGGT.
We sequentially determine the selection threshold $\epsilon$,
the Top-$K$ constraint, and the transformer blocks where
view selection is applied.
}
\label{tab:ablation}

\footnotesize
\setlength{\tabcolsep}{10pt}
\renewcommand{\arraystretch}{1.08}

\resizebox{0.5\textwidth}{!}{%
\begin{tabular}{ccc|ccccc}
\specialrule{1pt}{0pt}{2pt}
\specialrule{0.3pt}{0pt}{0pt}
Threshold $\epsilon$
& Top-$K$
& Blocks
& CD$\downarrow$
& ATE$\downarrow$
& ARE$\downarrow$
& RPE-t$\downarrow$
& RPE-r$\downarrow$ \\
\hline

\rowcolor{gray!15}
\multicolumn{3}{c|}{\textbf{Baseline}}
& 0.4375 & 0.0920 & 3.9553 & 0.0559 & 1.2580 \\

\hline

\cellcolor{ablred} 0.05
& \checkmark & 0--23
& 0.4223 & 0.0950 & 4.0080 & 0.0601 & 1.2739 \\

\cellcolor{ablred} 0.10
& \checkmark & 0--23
& 0.4147 & 0.0988 & 4.0749 & 0.0619 & 1.2715 \\

\cellcolor{ablred} 0.15
& \checkmark & 0--23
& 0.4123 & 0.0988 & 4.1532 & 0.0648 & 1.3128 \\

\cellcolor{ablred} \textbf{0.20}
& \checkmark & 0--23
& 0.4109 & 0.1032 & 4.1628 & 0.0667 & 1.3570 \\

\cellcolor{ablred} 0.25
& \checkmark & 0--23
& 0.4123 & 0.1134 & 4.2518 & 0.0737 & 1.4210 \\

\cellcolor{ablred} 0.30
& \checkmark & 0--23
& 0.4078 & 0.1160 & 4.4115 & 0.0787 & 1.5592 \\

\cellcolor{ablred} 0.35
& \checkmark & 0--23
& 0.4011 & 0.1244 & 4.5468 & 0.0872 & 1.7130 \\

\cellcolor{ablred} 0.40
& \checkmark & 0--23
& 0.4098 & 0.1477 & 5.7083 & 0.1062 & 2.3883 \\

\hline

0.20
& \cellcolor{ablyellow} \textbf{w/}
& 0--23
& 0.4109 & 0.1032 & 4.1628 & 0.0667 & 1.3570 \\

0.20
& \cellcolor{ablyellow} w/o
& 0--23
& 0.4113 & 0.1038 & 4.1659 & 0.0669 & 1.3669 \\

\hline

0.20
& \checkmark
& \cellcolor{ablgreen} \textbf{0--23}
& 0.4109 & 0.1032 & 4.1628 & 0.0667 & 1.3570 \\

0.20
& \checkmark
& \cellcolor{ablgreen} 8--23
& 0.4133 & 0.1068 & 4.2634 & 0.0679 & 1.3552 \\

0.20
& \checkmark
& \cellcolor{ablgreen} 12--23
& 0.4185 & 0.1081 & 4.2267 & 0.0713 & 1.4026 \\

0.20
& \checkmark
& \cellcolor{ablgreen} 18--23
& 0.4373 & 0.0923 & 3.9627 & 0.0559 & 1.1998 \\

\hline
\end{tabular}
}
\vspace{-0.6cm}
\end{table}

\begin{table*}
\centering
\caption{
Compatibility on ScanNet-50~\cite{dai2017scannet}: point cloud reconstruction and runtime
before and after adding BlazeVGGT into FastVGGT~\cite{shen2025fastvggt} and LiteVGGT~\cite{litevggt2025}..
}
\label{tab:compat_cd}

\footnotesize
\setlength{\tabcolsep}{7pt}
\renewcommand{\arraystretch}{1.25}

\resizebox{\textwidth}{!}{%
\begin{tabular}{l|cc|cc|cc|cc}
\specialrule{1pt}{0pt}{2pt}
\specialrule{0.3pt}{0pt}{0pt}
\multicolumn{1}{c|}{\multirow{2}{*}{Method}}
& \multicolumn{2}{c|}{1,000}
& \multicolumn{2}{c|}{500}
& \multicolumn{2}{c|}{300}
& \multicolumn{2}{c}{100} \\
\cline{2-9}
& CD$\downarrow$ & Time$\downarrow$
& CD$\downarrow$ & Time$\downarrow$
& CD$\downarrow$ & Time$\downarrow$
& CD$\downarrow$ & Time$\downarrow$ \\
\hline




FastVGGT (ICLR'26)
& 0.4654 & 180.1s
& 0.4569 & 58.0s
& 0.4467 & 24.5s
& 0.4234 & 4.8s \\

\rowcolor{accentcyan}
FastVGGT + BlazeVGGT
& \textbf{0.4255} & \textbf{81.6s} \textcolor{green!50!black}{(\scriptsize $\uparrow$2.2$\times$)}
& \textbf{0.4224} & \textbf{33.1s} \textcolor{green!50!black}{(\scriptsize $\uparrow$1.8$\times$)}
& \textbf{0.4238} & \textbf{15.8s} \textcolor{green!50!black}{(\scriptsize $\uparrow$1.6$\times$)}
& 0.4112 & \textbf{4.2s} \textcolor{green!50!black}{(\scriptsize $\uparrow$1.1$\times$)} \\

\hline

LiteVGGT (CVPR'26)
& 0.4085 & 133.8s
& 0.3862 & 44.2s
& 0.3516 & 24.7s
& \textbf{0.3291} & \textbf{8.1s} \\

\rowcolor{accentcyan}
LiteVGGT + BlazeVGGT
& \textbf{0.3640} & \textbf{45.8s} \textcolor{green!50!black}{(\scriptsize $\uparrow$2.9$\times$)}
& \textbf{0.3713} & \textbf{23.8s} \textcolor{green!50!black}{(\scriptsize $\uparrow$1.9$\times$)}
& \textbf{0.3397} & \textbf{15.5s} \textcolor{green!50!black}{(\scriptsize $\uparrow$1.6$\times$)}
& 0.3409 & 8.9s \textcolor{red!70!black}{(\scriptsize $\downarrow$1.1$\times$)} \\

\hline
\end{tabular}
}
\vspace{-0.2cm}
\end{table*}

\begin{table}
\centering
\caption{
Point cloud reconstruction on Tanks \& Temples~\cite{knapitsch2017tanks}.
We report scene-wise F1-score and average inference time.
$\times5$ and $\times10$ indicate evaluation thresholds scaled by
five and ten times the official values, respectively.
}
\label{tab:recon_tanks_temples}

\footnotesize
\setlength{\tabcolsep}{6pt}
\renewcommand{\arraystretch}{1.15}

\resizebox{0.5\textwidth}{!}{%
\begin{tabular}{l|cccccc|cc}
\specialrule{1pt}{0pt}{2pt}
\specialrule{0.3pt}{0pt}{0pt}
\multicolumn{1}{c|}{Method}
& Barn
& Caterpillar
& Courthouse
& Ignatius
& Meetingroom
& Truck
& Avg.$\uparrow$
& Time(s)$\downarrow$ \\
\specialrule{0.8pt}{0pt}{1pt}

\rowcolor{gray!15}
\multicolumn{1}{c|}{threshold}
& 0.010 & 0.005 & 0.025 & 0.003 & 0.010 & 0.005
& $\times5$ & -- \\
\specialrule{0.5pt}{1pt}{0pt}



FastVGGT\cite{shen2025fastvggt} (ICLR'26)
& 0.35 & 0.27 & 0.34 & 0.35 & 0.40 & 0.48 & \textbf{0.36} & 63.51s \\

\rowcolor{accentcyan}
FastVGGT + BlazeVGGT
& 0.29 & 0.29 & 0.34 & 0.31 & 0.44 & 0.52 & \textbf{0.36} & \textbf{41.92s} \textcolor{green!50!black}{(\scriptsize $\uparrow$1.5$\times$)} \\ \hline

LiteVGGT\cite{litevggt2025} (CVPR'26)
& 0.42 & 0.29 & 0.24 & 0.30 & 0.57 & 0.54 & \textbf{0.39} & 30.79s \\

LiteVGGT + BlazeVGGT
& 0.39 & 0.28 & 0.26 & 0.33 & 0.55 & 0.51 & \textbf{0.39} & \textbf{17.78s} \textcolor{green!50!black}{(\scriptsize $\uparrow$1.7$\times$)} \\ 

\specialrule{0.8pt}{0pt}{1pt}

\rowcolor{gray!15}
\multicolumn{1}{c|}{threshold}
& 0.010 & 0.005 & 0.025 & 0.003 & 0.010 & 0.005
& $\times10$ & -- \\
\specialrule{0.5pt}{1pt}{0pt}



FastVGGT\cite{shen2025fastvggt} (ICLR'26)
& 0.58 & 0.44 & 0.51 & 0.53 & 0.66 & 0.64 & \textbf{0.56} & 63.51s \\

\rowcolor{accentcyan}
FastVGGT + BlazeVGGT
& 0.50 & 0.47 & 0.51 & 0.47 & 0.69 & 0.67 & {0.55} & \textbf{41.92s} \textcolor{green!50!black}{(\scriptsize $\uparrow$1.5$\times$)} \\ \hline

LiteVGGT\cite{litevggt2025} (CVPR'26)
& 0.67 & 0.51 & 0.42 & 0.45 & 0.78 & 0.70 & \textbf{0.59} & 30.79s \\

\rowcolor{accentcyan}
LiteVGGT + BlazeVGGT
& 0.62 & 0.50 & 0.44 & 0.48 & 0.76 & 0.70 & {0.58} & \textbf{17.78s} \textcolor{green!50!black}{(\scriptsize $\uparrow$1.7$\times$)} \\ 

\bottomrule
\end{tabular}
}
\vspace{-0.3cm}
\end{table}

\begin{table*}
\centering
\caption{
Compatibility on ScanNet-50~\cite{dai2017scannet}: camera pose errors before and after
integrating BlazeVGGT into VGGT~\cite{wang2025vggt}, FastVGGT~\cite{shen2025fastvggt}, and LiteVGGT~\cite{litevggt2025}.
We report ATE, ARE, RPE-t, and RPE-r. Lower is better.
}
\label{tab:compat_pose}

\footnotesize
\setlength{\tabcolsep}{10pt}
\renewcommand{\arraystretch}{1.10}

\resizebox{\textwidth}{!}{%
\begin{tabular}{l|cccc|cccc|cccc|cccc}
\specialrule{1pt}{0pt}{2pt}
\specialrule{0.3pt}{0pt}{0pt}
\multicolumn{1}{c|}{\multirow{2}{*}{Method}}
& \multicolumn{4}{c|}{1,000}
& \multicolumn{4}{c|}{500}
& \multicolumn{4}{c|}{300}
& \multicolumn{4}{c}{100} \\
\cline{2-17}
& ATE$\downarrow$ & ARE$\downarrow$ & RPE-t$\downarrow$ & RPE-r$\downarrow$
& ATE$\downarrow$ & ARE$\downarrow$ & RPE-t$\downarrow$ & RPE-r$\downarrow$
& ATE$\downarrow$ & ARE$\downarrow$ & RPE-t$\downarrow$ & RPE-r$\downarrow$
& ATE$\downarrow$ & ARE$\downarrow$ & RPE-t$\downarrow$ & RPE-r$\downarrow$ \\
\hline




FastVGGT (ICLR'26)
& 0.115 & 4.014 & {0.031} & \textbf{0.663}
& \textbf{0.097} & \textbf{4.151} & \textbf{0.032} & \textbf{0.717}
& \textbf{0.092} & \textbf{4.033} & \textbf{0.036} & \textbf{0.762}
& \textbf{0.093} & \textbf{3.853} & \textbf{0.060} & \textbf{1.257} \\

\rowcolor{accentcyan}
FastVGGT + BlazeVGGT
& \textbf{0.099} & \textbf{3.732} & \textbf{0.029} & 0.683
& 0.106 & 4.430 & 0.035 & 0.805
& 0.107 & 4.542 & 0.044 & 0.986
& 0.111 & 4.362 & 0.075 & 1.492 \\

\hline

LiteVGGT (CVPR'26)
& 0.150 & 7.444 & 0.039 & 1.107
& \textbf{0.120} & \textbf{4.678} & 0.041 & 0.791
& \textbf{0.114} & 4.463 & \textbf{0.046} & \textbf{0.896}
& \textbf{0.108} & \textbf{4.111} & \textbf{0.073} & \textbf{1.448} \\

\rowcolor{accentcyan}
LiteVGGT + BlazeVGGT
& \textbf{0.108} & \textbf{4.249} & \textbf{0.032} & \textbf{0.630}
& \textbf{0.120} & 4.856 & \textbf{0.039} & \textbf{0.746}
& 0.117 & \textbf{4.409} & 0.048 & 0.903
& 0.123 & 4.335 & 0.082 & 1.650 \\

\hline
\end{tabular}
}
\end{table*}

\begin{table*}
\centering
\caption{
Online reconstruction with StreamVGGT~\cite{zhuo2025streaming} under increasing context length.
Each entry reports accuracy, inference time (ms), and memory consumption
(GB).
}
\label{tab:streamvggt_online}

\footnotesize
\renewcommand{\arraystretch}{1.15}

\resizebox{\textwidth}{!}{%
\begin{tabular}{l|cccccc}
\specialrule{1pt}{0pt}{2pt}
\specialrule{0.3pt}{0pt}{0pt}
\multicolumn{1}{c|}{\multirow{2}{*}{Method}}
& 1 frame
& 5 frames
& 10 frames
& 20 frames
& 30 frames
& 40 frames \\
& Acc$\downarrow$ / ms$\downarrow$ / GB$\downarrow$
& Acc$\downarrow$ / ms$\downarrow$ / GB$\downarrow$
& Acc$\downarrow$ / ms$\downarrow$ / GB$\downarrow$
& Acc$\downarrow$ / ms$\downarrow$ / GB$\downarrow$
& Acc$\downarrow$ / ms$\downarrow$ / GB$\downarrow$
& Acc$\downarrow$ / ms$\downarrow$ / GB$\downarrow$ \\
\hline

Spann3R\cite{wang20243d} (3DV'25)
& 0.039 / 500 / 5.3
& 0.038 / \textbf{87} / 5.4
& 0.046 / \textbf{82} / 5.5
& 0.036 / 120 / 5.5
& 0.043 / 192 / 5.6
& 0.068 / 125 / 5.7 \\

CUT3R\cite{wang2025continuous} (CVPR'25)
& 0.035 / 101 / 3.4
& 0.038 / 102 / 3.5
& 0.039 / 99 / 3.6
& 0.031 / \textbf{101} / \textbf{3.8}
& 0.027 / 102 / \textbf{4.0}
& 0.023 / 102 / \textbf{4.2} \\

StreamVGGT (ICLR'26)
& 0.028 / \textbf{63} / \textbf{2.1}
& 0.024 / 88 / \textbf{2.7}
& 0.024 / 120 / \textbf{3.2}
& 0.021 / 152 / 4.3
& 0.021 / 184 / 5.5
& 0.021 / 216 / 6.6 \\

\rowcolor{accentcyan}
StreamVGGT + BlazeVGGT
& \textbf{0.021} / 96 / 7.2
& \textbf{0.017} / 96 / 7.5
& \textbf{0.015} / 97 / 7.5
& \textbf{0.014} / 107 / 7.7
& \textbf{0.014} / \textbf{96} / 7.8
& \textbf{0.015} / \textbf{97} / 8.0 \\

\hline
\end{tabular}%
}
\end{table*}

\begin{table*}
\centering
\caption{
Comparison of context-pruning strategies for StreamVGGT~\cite{zhuo2025streaming} on a
200-frame online sequence.
}
\label{tab:stream_pruning}

\footnotesize
\setlength{\tabcolsep}{8pt}
\renewcommand{\arraystretch}{1.2}

\resizebox{\textwidth}{!}{%
\begin{tabular}{ll|cc|cc|cc|cc}
\specialrule{1pt}{0pt}{2pt}
\specialrule{0.3pt}{0pt}{0pt}
\multicolumn{2}{c|}{\multirow{2}{*}{Method}}
& \multicolumn{2}{c|}{Acc$\downarrow$}
& \multicolumn{2}{c|}{Comp$\downarrow$}
& \multicolumn{2}{c|}{NC$\uparrow$}
& \multirow{2}{*}{\shortstack{Inference Time\\@ frame 200}}
& \multirow{2}{*}{Peak Memory} \\
\cline{3-8}
& & Mean & Med. & Mean & Med. & Mean & Med. & & \\
\hline

\multicolumn{2}{l|}{Spann3R\cite{wang20243d} (3DV'25)}
& 0.225 & 0.176
& 0.070 & 0.015
& 0.538 & 0.557
& 221.054 ms & 6.7 GB \\

\multicolumn{2}{l|}{CUT3R\cite{wang2025continuous} (CVPR'25)}
& 0.165 & 0.107
& 0.030 & 0.012
& 0.551 & 0.577
& 102.533 ms & 7.6 GB \\

\hline

\multirow{6}{*}{StreamVGGT (ICLR'26)}
& None
& \textbf{0.030} & \uline{0.024}
& \uline{0.021} & \uline{0.013}
& \uline{0.560} & \uline{0.561}
& 737.535 ms & 31.9 GB \\

& Window(50)
& 0.062 & 0.033
& 0.028 & 0.015
& 0.553 & 0.554
& 277.584 ms & 20.3 GB \\

& Window(100)
& 0.043 & 0.028
& 0.023 & 0.014
& 0.557 & 0.557
& 478.251 ms & 30.7 GB \\

& K-nearest(50)
& 0.062 & 0.032
& 0.028 & 0.015
& 0.555 & 0.554
& \uline{274.241 ms} & \uline{20.1 GB} \\

& K-nearest(100)
& 0.041 & 0.028
& 0.023 & 0.014
& 0.557 & 0.557
& 475.945 ms & 30.5 GB \\

\rowcolor{accentcyan}
& BlazeVGGT
& \uline{0.034} & \textbf{0.023}
& \textbf{0.016} & \textbf{0.012}
& \textbf{0.567} & \textbf{0.565}
& \textbf{94.979 ms} & \textbf{10.2 GB} \\

\hline
\end{tabular}

}
\vspace{-0.3cm}
\end{table*}

\noindent{\textit{(\romannumeral 2) Camera Pose Estimation.}}
Having established the reconstruction efficiency of BlazeVGGT, we
further examine whether its compact KV context preserves the global
geometric reasoning required for camera pose estimation.
TABLE~\ref{tab:scannet_pose} shows that BlazeVGGT remains highly
competitive across different sequence lengths. Most notably, at
1,000 views, it achieves the lowest ATE, ARE, RPE-t, and RPE-r among
all compared methods, indicating that a compact set of complementary
KV views can retain sufficient global context for accurate trajectory
estimation. Compared with the trained variants, BlazeVGGT outperforms
both LiteVGGT and VGG-T$^3$ across all four pose metrics at 500 and
1,000 views, despite requiring neither additional training nor
input-specific optimization. Within the more directly comparable
training-free group, BlazeVGGT achieves the best overall pose accuracy
at both 1,000 and 2,000 views and remains close to FastVGGT at
500 views. 
Taken together with the substantial runtime reductions reported above,
these results show that BlazeVGGT can preserve competitive global pose
reasoning while greatly improving long-sequence inference efficiency.

\noindent\textbf{\textit{Qualitative Comparison.}}
We complement the quantitative evaluation with visual comparisons of
reconstructed point clouds and estimated camera trajectories.

\begin{figure*}[p]
\centering
\captionsetup[subfigure]{font=footnotesize,skip=2pt}
\begin{subfigure}{0.97\textwidth}
\centering
\includegraphics[width=\linewidth]{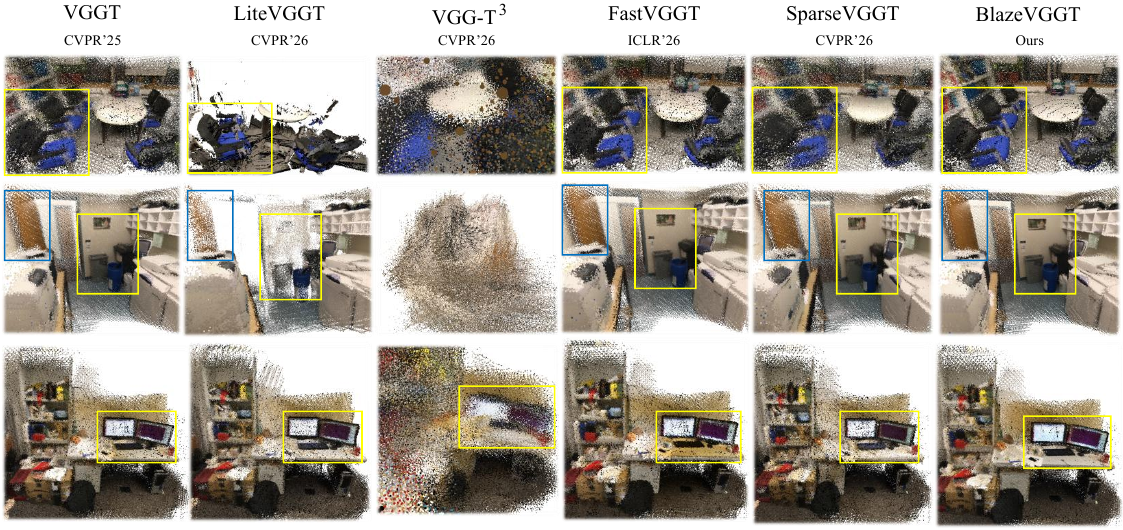}
\caption{Local reconstruction quality.}
\label{fig:point_cloud_a}
\end{subfigure}
\par\vspace{4pt}
\begin{subfigure}{0.97\textwidth}
\centering
\includegraphics[width=\linewidth]{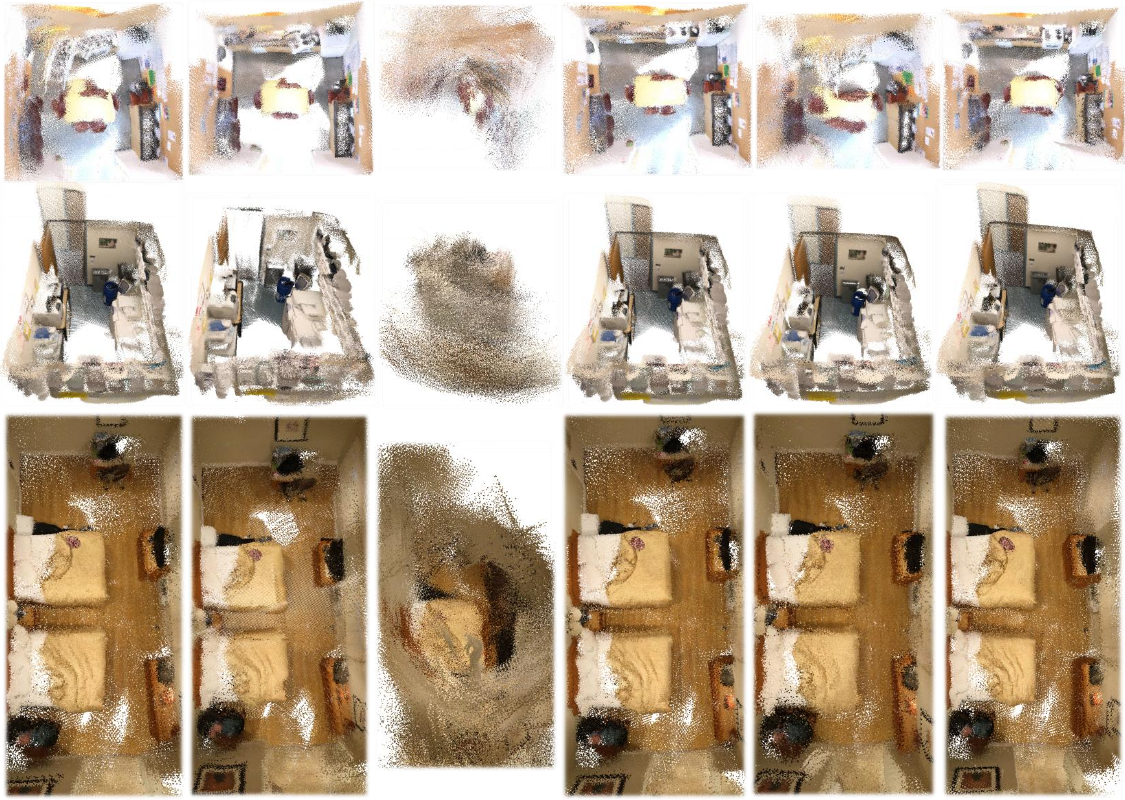}
\caption{Full-scene completeness.}
\label{fig:point_cloud_b}
\end{subfigure}
\caption{Qualitative comparison of point cloud reconstruction. From left
to right: VGGT~\cite{wang2025vggt}, LiteVGGT~\cite{litevggt2025}, VGG-T$^3$~\cite{vggttt2026}, FastVGGT~\cite{shen2025fastvggt}, SparseVGGT~\cite{sparsevggt2025}, and
BlazeVGGT (Ours). (a) Selected viewpoints within
the full reconstructed scenes; colored rectangles highlight local
regions that demonstrate BlazeVGGT's reconstruction quality.
(b) Full scenes, demonstrating that BlazeVGGT
preserves scene completeness and coherent global structure despite
aggressive frame-level pruning of the KV context.}
\label{fig:example_point_comparison}
\end{figure*}

\noindent{\textit{(\romannumeral 1) Point Cloud Reconstruction.}}
Fig.~\ref{fig:example_point_comparison} highlights BlazeVGGT's strong
reconstruction quality at both the local and scene levels.
Fig.~\ref{fig:example_point_comparison}(\subref{fig:point_cloud_a}) presents selected viewpoints within the full reconstructed
scenes, with colored rectangles drawing attention to regions where
BlazeVGGT excels in recovering fine geometric details. In the first
row, BlazeVGGT reconstructs distinct chair backs and seats with clear
separation from the surrounding furniture, whereas LiteVGGT suffers
from severe fragmentation, VGG-T$^3$ produces a diffuse point cloud,
and SparseVGGT blurs the chair boundaries. In the second row, the
blue and yellow rectangles highlight the well-defined doorway and
the coherent wall and objects at the back of the room. BlazeVGGT
preserves these structures clearly, while LiteVGGT leaves substantial
gaps and VGG-T$^3$ fails to recover a recognizable room layout.
In the third-row yellow boxes, BlazeVGGT reconstructs a distinct keyboard,
a coherent desktop surface, and sharp monitor boundaries, whereas
others smears these details. Together, these local
views demonstrate that BlazeVGGT retains crisp object boundaries and
spatially coherent details even under substantial sparsification.
Fig.~\ref{fig:example_point_comparison}(\subref{fig:point_cloud_b}) shifts the focus to full-scene reconstruction,
showing that aggressive frame-level pruning of the KV context does
not come at the expense of scene completeness. BlazeVGGT retains
the enclosing room structure, broad floor coverage, and the spatial
arrangement of furniture across these examples. Its reconstructions
avoid the severe structural collapse of VGG-T$^3$ and the pronounced
distortion of SparseVGGT in the first row of (b), while preserving the
scene coverage visible in the dense VGGT baseline. This combination
of detailed local geometry and complete scene structure is a central
advantage of BlazeVGGT: by retaining all query frames and selecting
complementary KV views, it achieves substantial computational savings
while delivering visually compelling reconstructions at both scales.

\noindent{\textit{(\romannumeral 2) Camera Pose Estimation.}}
Fig.~\ref{fig:camera_pose_visualization} provides qualitative comparisons of the estimated camera trajectories. In the more challenging first sequence, BlazeVGGT has the lowest displayed ATE and ARE and visibly reduces trajectory drift. In the second sequence, BlazeVGGT remains closely aligned with the ground truth and improves on VGGT and LiteVGGT, while FastVGGT has a slightly lower displayed ATE. These visualizations are consistent with the metric-dependent trends in TABLE~\ref{tab:scannet_pose}.

\subsection{Ablation Studies}
\label{sec:ablation}

TABLE~\ref{tab:ablation} analyzes three design choices of BlazeVGGT: the diversity threshold $\epsilon$, the Top-$K$ constraint, and the range of global-attention blocks to which view selection is applied.

\noindent\textbf{\textit{Selection Threshold.}} The diversity threshold $\epsilon$ controls the aggressiveness of determinantal view selection and therefore the trade-off between reconstruction and pose metrics. Relative to the baseline, $\epsilon=0.20$ reduces CD from 0.4375 to 0.4109, while increasing ATE from 0.0920 to 0.1032. The lowest CD in this ablation is obtained at $\epsilon=0.35$ (0.4011), but its ATE, ARE, RPE-t, and RPE-r are all higher than at $\epsilon=0.20$. Increasing $\epsilon$ beyond 0.20 does not monotonically improve CD: it worsens at $\epsilon=0.25$ and again at $\epsilon=0.40$ relative to their preceding settings, while the pose errors increase overall. We therefore use $\epsilon=0.20$ as a balanced default rather than as the optimum of any individual metric.

\noindent\textbf{\textit{Top-$K$ Constraint.}} We next study whether the explicit upper bound on the number of selected views affects prediction quality. Enabling or removing Top-$K$ results in nearly identical CD and camera-pose metrics, showing that the performance gains of BlazeVGGT primarily arise from the diversity-driven selection criterion rather than from a fixed cardinality constraint. The role of Top-$K$ is therefore mainly practical: it acts as a safety ceiling for blocks in which the adaptive criterion would otherwise retain an unusually large KV set, while the effective number of selected views remains determined by the camera-token similarity structure. We retain this constraint in the default configuration to make the computational budget more predictable without materially altering reconstruction or pose accuracy.

\noindent\textbf{\textit{Layer Range.}} Finally, we vary the global-attention blocks in which BlazeVGGT is activated. Applying view selection only to the deepest blocks leaves the reconstruction result close to the unsparsified baseline, whereas progressively extending sparsification toward earlier blocks consistently improves CD. This observation is closely aligned with the layer-wise analysis in Fig.~\ref{fig:motivation_rsa}. Although shallow camera tokens do not yet encode accurately calibrated global geometry, their local relational structure is already sufficient to distinguish nearby and redundant views. Consequently, useful redundancy suppression does not need to wait until explicit global geometry has emerged. 
Restricting selection to late layers preserves camera-pose accuracy more closely but leaves a substantial portion of redundant global-attention computation untouched. In contrast, applying BlazeVGGT throughout the stack yields the strongest reconstruction improvement and suppresses redundant KV computation at every stage, thereby offering greater acceleration potential.
We therefore apply view selection to all global-attention blocks in the default configuration.

\vspace{-0.2cm}
\subsection{Compatibility with Existing VGGT Accelerators}
\label{sec:compatibility}

By targeting view-level redundancy without modifying pretrained weights, BlazeVGGT is complementary to other VGGT acceleration mechanisms. We evaluate this property by directly combining it with FastVGGT~\cite{shen2025fastvggt} and LiteVGGT~\cite{litevggt2025}, without retraining.

\noindent\textbf{\textit{Point Cloud Reconstruction.}} TABLES~\ref{tab:compat_cd} and~\ref{tab:recon_tanks_temples} demonstrate consistent compositional behavior on both ScanNet-50 and Tanks \& Temples~\cite{knapitsch2017tanks}. On ScanNet-50, adding BlazeVGGT to either FastVGGT or LiteVGGT provides substantial additional acceleration in the long-sequence regime while generally improving reconstruction quality. The gains are particularly clear at 1,000 frames, where both composed variants achieve lower CD and considerably shorter runtime than their corresponding backbones. 
The same overall behavior transfers to Tanks \& Temples. BlazeVGGT provides further speedups when composed with both FastVGGT and LiteVGGT while leaving their average F1 scores nearly unchanged across the two evaluation thresholds.  The consistency across ScanNet-50 and Tanks \& Temples indicates that frame-level sparsification can serve as a plug-and-play complement to existing VGGT acceleration mechanisms rather than being tied to a particular accelerated backbone.

\noindent\textbf{\textit{Camera Pose Estimation.}} TABLE~\ref{tab:compat_pose} shows a similar long-context trend for camera pose estimation on ScanNet-50. At 1,000 frames, integrating BlazeVGGT improves the absolute pose errors of FastVGGT and improves all reported pose metrics for LiteVGGT, while substantially reducing their reconstruction-time cost as shown above. For shorter sequences, the additional sparsification can introduce modest degradation on some pose metrics, particularly when the underlying accelerator has already compressed the representation. These results suggest that different acceleration mechanisms provide largely complementary benefits in the target regime of BlazeVGGT, namely long image sequences with substantial redundancy. When the input context is already small, however, combining multiple aggressive acceleration strategies does not necessarily produce consistent improvements across all metrics.

\vspace{-0.2cm}
\subsection{Generalization to VGGT-based Models}
\label{sec:generalization}

Beyond its compatibility with VGGT accelerators, we evaluate the transferability of view-level sparsification to StreamVGGT for streaming reconstruction and PAGE-4D for video geometry. In both settings, BlazeVGGT is integrated without retraining the underlying model.

\begin{figure}
\centering
\includegraphics[width=\columnwidth]{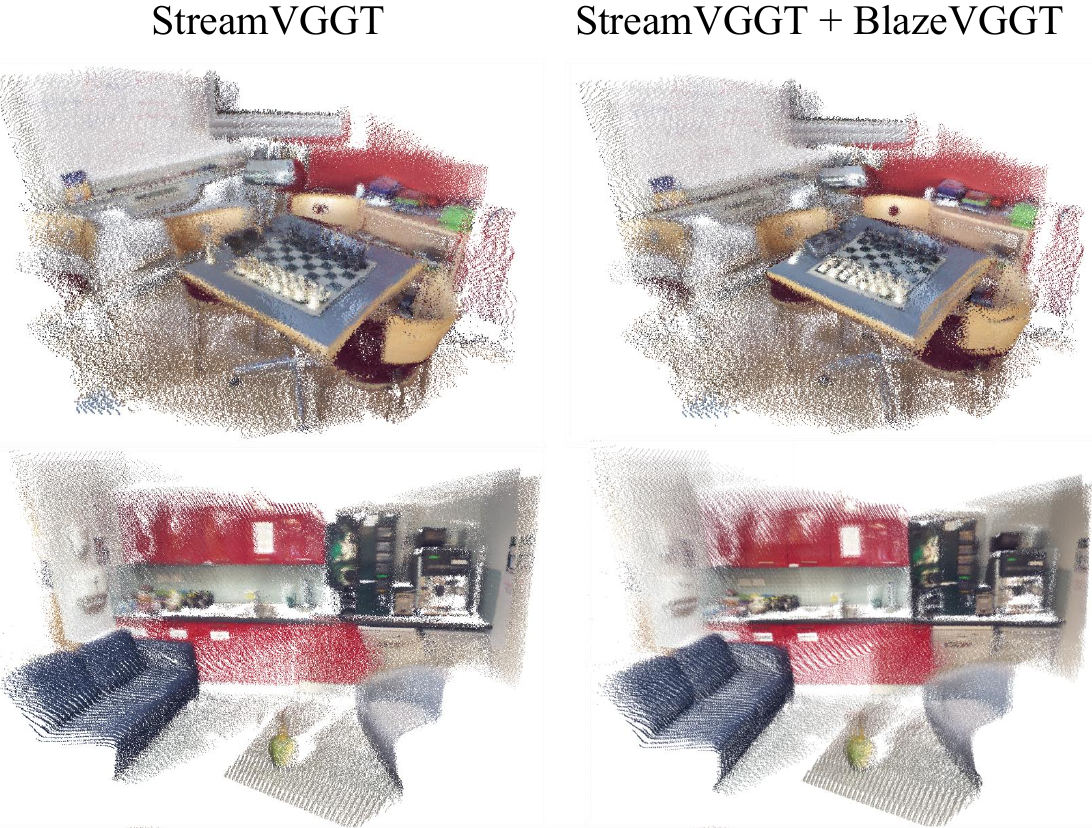}
\caption{Qualitative point-cloud comparison for online reconstruction with StreamVGGT~\cite{zhuo2025streaming}. Across two scenes, StreamVGGT~\cite{zhuo2025streaming} with BlazeVGGT preserves coherent scene layouts and object boundaries while using a compact long-term context.}
\label{fig:streamvggt_point_cloud_cmp}
\vspace{-0.1cm}
\end{figure}

\noindent\textbf{\textit{Online Reconstruction with StreamVGGT.}} The kernel scale in the original BlazeVGGT depends on the median of pairwise descriptor distances over the entire sequence. Since single-pass streaming inference cannot access future observations at the current time step, this full-sequence statistic cannot be reproduced exactly. To establish a streaming baseline that does not rely on scene priors, we fix the kernel scale to its default value; all StreamVGGT + BlazeVGGT experiments and visualizations in this paper use this setting.
Building on this baseline, we recommend adopting a dual-memory design: a short-term FIFO memory retains recent observations and local temporal context, while a determinantal criterion maintains the long-term KV history. When a frame leaves the FIFO memory and is considered for insertion into long-term memory, the kernel scale is computed online using only the descriptors of the current long-term memory and the candidate frame. This allows the kernel to adapt to the structure and redundancy of the scene observed so far, without access to future frames or priors about the entire scene.
TABLE~\ref{tab:streamvggt_online} shows that the latency of the original StreamVGGT increases steadily as its historical context grows, whereas StreamVGGT with BlazeVGGT maintains nearly constant inference time after the context becomes sufficiently long, while also achieving consistently strong reconstruction accuracy. The advantage becomes clearer in the 200-frame evaluation of TABLE~\ref{tab:stream_pruning}. Compared with the unpruned StreamVGGT, determinantal selection reduces the per-frame latency from roughly 738\,ms to 95\,ms and peak memory from 31.9\,GB to 10.2\,GB, while remaining close in accuracy and improving completeness and normal consistency. It also outperforms fixed-window and nearest-neighbor pruning in both efficiency and memory usage. Fig.~\ref{fig:streamvggt_point_cloud_cmp} provides complementary point-cloud comparisons on two online scenes. The BlazeVGGT variant preserves coherent room layouts and clearer object boundaries while maintaining a compact long-term context. This result suggests that geometric diversity, rather than temporal recency or independently selected nearest views, provides a more effective criterion for maintaining a compact long-term context in streaming visual geometry.

\begin{table}[t]
\centering
\caption{
Video depth estimation on Sintel~\cite{Butler:ECCV:2012}, Bonn~\cite{palazzolo2019iros}, and DyCheck~\cite{gao2022dynamic} under standard and
extended sequence lengths. We report Abs Rel, $\delta<1.25$, and
inference throughput.
}
\label{tab:depth_estimation}

\footnotesize
\setlength{\tabcolsep}{3pt}
\renewcommand{\arraystretch}{1.05}

\resizebox{\columnwidth}{!}{%
\begin{tabular}{lc|cc|cc|cc|c}
\specialrule{1pt}{0pt}{2pt}
\specialrule{0.3pt}{0pt}{0pt}
\multicolumn{1}{c}{\multirow{2}{*}{Method}}
& \multirow{2}{*}{Params}
& \multicolumn{2}{c|}{Sintel\cite{Butler:ECCV:2012}}
& \multicolumn{2}{c|}{Bonn\cite{palazzolo2019iros}}
& \multicolumn{2}{c|}{DyCheck\cite{gao2022dynamic}}
& \multirow{2}{*}{FPS} \\
\cline{3-8}
&
&
\cellcolor{absrel}{Abs Rel$\downarrow$}
& \cellcolor{delta}{$\delta<1.25\uparrow$}
& \cellcolor{absrel}{Abs Rel$\downarrow$}
& \cellcolor{delta}{$\delta<1.25\uparrow$}
& \cellcolor{absrel}{Abs Rel$\downarrow$}
& \cellcolor{delta}{$\delta<1.25\uparrow$}
& \\
\specialrule{0.8pt}{0pt}{0pt}
\addlinespace[1pt]

\rowcolor{framesetting}
\multicolumn{9}{c}{Around 100 Frames} \\

\specialrule{0.5pt}{1pt}{1pt}

\rowcolor{alignsetting}
\multicolumn{9}{l}{\textit{Depth alignment}} \\
\addlinespace[1pt]

PAGE4D\cite{ICLR2026_3d551707} (ICLR'26)
& 1.26B
& 0.357 & 0.699
& 0.022 & 0.723
& 0.205 & 0.767
& 11.18 \\

\rowcolor{ours}
PAGE4D + BlazeVGGT
& 1.26B
& \textbf{0.312} & \textbf{0.703}
& \textbf{0.088} & \textbf{0.911}
& \textbf{0.204} & \textbf{0.774}
& \textbf{21.31} \\

\specialrule{0.5pt}{0pt}{1pt}

\rowcolor{alignsetting}
\multicolumn{9}{l}{\textit{Scale-and-shift alignment}} \\

\addlinespace[1pt]

PAGE4D\cite{ICLR2026_3d551707} (ICLR'26)
& 1.26B
& 0.212 & \textbf{0.763}
& 0.090 & 0.903
& \textbf{0.183} & \textbf{0.802}
& 11.18 \\

\rowcolor{ours}
PAGE4D + BlazeVGGT
& 1.26B
& \textbf{0.206} & {0.751}
& \textbf{0.087} & \textbf{0.908}
& 0.185 & 0.799
& \textbf{21.31} \\

\specialrule{0.8pt}{0pt}{0pt}
\addlinespace[1pt]

\rowcolor{framesetting}
\multicolumn{9}{c}{Extended to 500 Frames} \\

\specialrule{0.5pt}{1pt}{1pt}

\rowcolor{alignsetting}
\multicolumn{9}{l}{\textit{Depth alignment}} \\

\addlinespace[1pt]

PAGE4D\cite{ICLR2026_3d551707} (ICLR'26)
& 1.26B
& -- & --
& \textbf{0.221} & \textbf{0.724}
& 0.210 & 0.721
& 0.81 \\

\rowcolor{ours}
PAGE4D + BlazeVGGT
& 1.26B
& -- & --
& 0.258 & 0.715
& \textbf{0.204} & \textbf{0.732}
& \textbf{5.06} \\

\specialrule{0.5pt}{0pt}{1pt}

\rowcolor{alignsetting}
\multicolumn{9}{l}{\textit{Scale-and-shift alignment}} \\
\addlinespace[1pt]

PAGE4D\cite{ICLR2026_3d551707} (ICLR'26)
& 1.26B
& -- & --
& \textbf{0.238} & \textbf{0.699}
& 0.200 & 0.736
& 0.81 \\

\rowcolor{ours}
PAGE4D + BlazeVGGT
& 1.26B
& -- & --
& 0.401 & 0.676
& \textbf{0.195} & \textbf{0.749}
& \textbf{5.06} \\

\hline
\end{tabular}
}
\end{table}

\noindent\textbf{\textit{Video Depth Estimation with PAGE-4D.}} We further integrate BlazeVGGT into PAGE-4D and evaluate video depth estimation on Sintel~\cite{Butler:ECCV:2012}, Bonn~\cite{palazzolo2019iros}, and DyCheck~\cite{gao2022dynamic}. As shown in TABLE~\ref{tab:depth_estimation}, BlazeVGGT nearly doubles the throughput around 100 frames while preserving broadly comparable depth estimation quality across the three benchmarks and the two alignment protocols. More importantly, the computational advantage grows rapidly with sequence length: at 500 frames, throughput increases from 0.81 FPS to 5.06 FPS. The accuracy trade-off at this scale is dataset dependent---DyCheck remains highly stable and even improves on the reported metrics, whereas Bonn exhibits a larger degradation under aggressive long-context sparsification. Thus, the transfer does not produce uniformly monotonic accuracy changes on every benchmark, but it consistently demonstrates the central scalability property of BlazeVGGT: as the input video becomes longer, a compact view-level KV context can substantially reduce the cost of a different VGGT-based geometry pipeline without modifying its learned parameters.

\section{Conclusion}

In this paper, we identify frame-level redundancy as a key bottleneck in VGGT and introduce BlazeVGGT, a training-free method that uses evolving camera-token relations to select complementary views for KV sparsification while retaining all queries. Experiments show substantial efficiency gains on long sequences with competitive geometric accuracy, alongside compatibility with existing accelerators and VGGT-based models. Future work will explore joint view-token sparsification and scene-adaptive computation.


%



\ifCLASSOPTIONcaptionsoff
  \newpage
\fi



\bibliographystyle{IEEEtran}
%

\bibliography{references}

@IEEEtranBSTCTL{BSTcontrol,
  CTLdash_repeated_names = "no"
}

@inproceedings{dust3r2024,
  title={{DUSt3R: Geometric 3D Vision Made Easy}},
  author={Wang, Shuzhe and Leroy, Vincent and Cabon, Yohann and Chidlovskii, Boris and Revaud, Jerome},
  booktitle={Proceedings of the IEEE/CVF Conference on Computer Vision and Pattern Recognition},
  year={2024}
}

@inproceedings{mast3r2025,
  title={{Grounding Image Matching in 3D with MASt3R}},
  author={Leroy, Vincent and Cabon, Yohann and Revaud, Jerome},
  booktitle={Proceedings of the IEEE/CVF Conference on Computer Vision and Pattern Recognition},
  year={2025}
}

@inproceedings{wang2025vggt,
  title={{VGGT: Visual Geometry Grounded Transformer}},
  author={Wang, Jianyuan and Chen, Minghao and Karaev, Nikita and Vedaldi, Andrea and Rupprecht, Christian and Novotny, David},
  booktitle={Proceedings of the IEEE/CVF Conference on Computer Vision and Pattern Recognition},
  pages={5294--5306},
  year={2025}
}

@article{vggtlong2025,
  title={{VGGT-Long: Chunk it, Loop it, Align it, Pushing VGGT's Limits on Kilometer-scale Long RGB Sequences}},
  author={Deng, Kai and Ti, Zexin and Xu, Jiawei and Yang, Jian and Xie, Jin},
  journal={arXiv preprint arXiv:2507.16443},
  year={2025}
}

@InProceedings{vggttt2026,
    author    = {Elflein, Sven and Li, Ruilong and Agostinho, S\'ergio and Gojcic, Zan and Leal-Taix\'e, Laura and Zhou, Qunjie and Osep, Aljosa},
    title     = {{VGG-T\${\textasciicircum}3\$: Offline Feed-Forward 3D Reconstruction at Scale}},
    booktitle = {Proceedings of the IEEE/CVF Conference on Computer Vision and Pattern Recognition},
    month     = {June},
    year      = {2026},
    pages     = {36464-36474}
}

@InProceedings{shen2025fastvggt,
  title={Fast{VGGT}: Fast Visual Geometry Transformer},
  author={You Shen and Zhipeng Zhang and Yansong Qu and Xiawu Zheng and Jiayi Ji and Shengchuan Zhang and Liujuan Cao},
  booktitle={International Conference on Learning Representations},
  year={2026},
}

@InProceedings{litevggt2025,
    author    = {Shu, Zhijian and Lin, Cheng and Xie, Tao and Yin, Wei and Li, Ben and Pu, Zhiyuan and Li, Weize and Yao, Yao and Cao, Xun and Guo, Xiaoyang and Long, Xiao-Xiao},
    title     = {{LiteVGGT: Boosting Vanilla VGGT via Geometry-aware Cached Token Merging}},
    booktitle = {Proceedings of the IEEE/CVF Conference on Computer Vision and Pattern Recognition},
    month     = {June},
    year      = {2026},
    pages     = {36422-36432}
}

@InProceedings{come2025,
    author    = {Chen, Yutian and Qiu, Yuheng and Li, Ruogu and Patrikar, Jay and Scherer, Sebastian},
    title     = {{Co-Me: Confidence Guided Token Merging for Visual Geometric Transformers}},
    booktitle = {Proceedings of the IEEE/CVF Conference on Computer Vision and Pattern Recognition},
    month     = {June},
    year      = {2026},
    pages     = {14590-14599}
}

@article{merg3r2026,
  title={{MERG3R: A Divide-and-Conquer Approach to Large-Scale Neural Visual Geometry}},
  author={Cheng, Leo Kaixuan and Shaikh, Abdus and Liang, Ruofan and Wu, Zhijie and Guan, Yushi and Vijaykumar, Nandita},
  journal={arXiv preprint arXiv:2603.02351},
  year={2026}
}

@InProceedings{httm2025,
    author    = {Wang, Weitian and Meiner, Lukas and Shubham, Rai and De La Parra, Cecilia and Kumar, Akash},
    title     = {{HTTM: Head-wise Temporal Token Merging for Faster VGGT}},
    booktitle = {Proceedings of the IEEE/CVF Conference on Computer Vision and Pattern Recognition},
    month     = {June},
    year      = {2026},
    pages     = {26379-26388}
}

@InProceedings{sparsevggt2025,
    author    = {Wang, Chung-Shien Brian and Schmidt, Christian and Piekenbrinck, Jens and Leibe, Bastian},
    title     = {{Block-Sparse Global Attention for Efficient Multi-View Geometry Transformers}},
    booktitle = {Proceedings of the IEEE/CVF Conference on Computer Vision and Pattern Recognition},
    month     = {June},
    year      = {2026},
    pages     = {14546-14555}
}

@InProceedings{flashvggt2025,
    author    = {Wang, Zipeng and Xu, Dan},
    title     = {{FlashVGGT: Efficient and Scalable Visual Geometry Transformers with Compressed Descriptor Attention}},
    booktitle = {Proceedings of the IEEE/CVF Conference on Computer Vision and Pattern Recognition},
    month     = {June},
    year      = {2026},
    pages     = {21826-21835}
}

@InProceedings{hess2026,
    author    = {Kim, Yongsung and Song, Wooseok and Lew, Jaihyun and Hwangbo, Hun and Lee, Jaehoon and Yoon, Sungroh},
    title     = {{HeSS: Head Sensitivity Score for Sparsity Redistribution in VGGT}},
    booktitle = {Proceedings of the IEEE/CVF Conference on Computer Vision and Pattern Recognition},
    month     = {June},
    year      = {2026},
    pages     = {36509-36517}
}

@InProceedings{avggt2025,
    author    = {Sun, Xianbing and Zhu, Zhikai and Lou, Zhengyu and Yang, Bo and Tang, Jinyang and Zhang, Liqing and Wang, He and Zhang, Jianfu},
    title     = {{AVGGT: Rethinking Global Attention for Accelerating VGGT}},
    booktitle = {Proceedings of the IEEE/CVF Conference on Computer Vision and Pattern Recognition},
    month     = {June},
    year      = {2026},
    pages     = {251-260}
}

@InProceedings{incvggt2026,
  author = {Fang, Keyu and Zhou, Changchun and Fu, Yuzhe and Li, Hai and Chen, Yiran },
  booktitle = {International Conference on Learning Representations},
  pages = {5609--5623},
  title = {{IncVGGT: Incremental VGGT for Memory-Bounded Long-Range 3D Reconstruction}},
  year = {2026}
}

@article{Kulesza_2012,
    author = {Kulesza, Alex and Taskar, Ben},
    title = {{Determinantal Point Processes for Machine Learning}},
    journal = {Foundations and Trends in Machine Learning},
    volume = {5},
    number = {2-3},
    pages = {123-286},
    year = {2012},
    month = {12},
}

@inproceedings{chen2018fastgreedymapinference,
 author = {Chen, Laming and Zhang, Guoxin and Zhou, Eric},
 booktitle = {Advances in Neural Information Processing Systems},
 pages = {},
 title = {{Fast Greedy MAP Inference for Determinantal Point Process to Improve Recommendation Diversity}},
 volume = {31},
 year = {2018}
}

@InProceedings{yang2025fast3r,
    author    = {Yang, Jianing and Sax, Alexander and Liang, Kevin J. and Henaff, Mikael and Tang, Hao and Cao, Ang and Chai, Joyce and Meier, Franziska and Feiszli, Matt},
    title     = {{Fast3R: Towards 3D Reconstruction of 1000+ Images in One Forward Pass}},
    booktitle = {Proceedings of the IEEE/CVF Conference on Computer Vision and Pattern Recognition},
    month     = {June},
    year      = {2025},
    pages     = {21924-21935}
}

@inproceedings{zhuo2025streaming,
 author = {Zhuo, Dong and Zheng, Wenzhao and Guo, Jiahe and Wu, Yuqi and Zhou, Jie and Lu, Jiwen},
 booktitle = {International Conference on Learning Representations},
 pages = {88055--88072},
 title = {Streaming Visual Geometry Transformer},
 year = {2026}
}

@InProceedings{wang2025continuous,
    author    = {Wang, Qianqian and Zhang, Yifei and Holynski, Aleksander and Efros, Alexei A. and Kanazawa, Angjoo},
    title     = {Continuous 3D Perception Model with Persistent State},
    booktitle = {Proceedings of the IEEE/CVF Conference on Computer Vision and Pattern Recognition},
    month     = {June},
    year      = {2025},
    pages     = {10510-10522}
}

@article{knapitsch2017tanks,
  author = {Knapitsch, Arno and Park, Jaesik and Zhou, Qian-Yi and Koltun, Vladlen},
  title = {Tanks and temples: benchmarking large-scale scene reconstruction},
  year = {2017},
  volume = {36},
  number = {4},
  journal = {ACM Transactions on Graphics},
  month = jul,
}

@INPROCEEDINGS{wang20243d,
  author={Wang, Hengyi and Agapito, Lourdes},
  booktitle={2025 International Conference on 3D Vision}, 
  title={3D Reconstruction with Spatial Memory}, 
  year={2025},
  volume={},
  number={},
  pages={78-89},
}

@book{hartley_multiple_2000,
  author = {Hartley, Richard and Zisserman, Andrew},
  title = {Multiple View Geometry in Computer Vision},
  year = {2003},
  publisher = {Cambridge University Press},
  edition = {2},
}

@inproceedings{snavely2006photo,
  author = {Snavely, Noah and Seitz, Steven M. and Szeliski, Richard},
  title = {Photo tourism: exploring photo collections in 3D},
  year = {2006},
  booktitle = {ACM SIGGRAPH},
  pages = {835–846},
}

@InProceedings{frahm2010building,
  author="Frahm, Jan-Michael
  and Fite-Georgel, Pierre
  and Gallup, David
  and Johnson, Tim
  and Raguram, Rahul
  and Wu, Changchang
  and Jen, Yi-Hung
  and Dunn, Enrique
  and Clipp, Brian
  and Lazebnik, Svetlana
  and Pollefeys, Marc",
  title="Building Rome on a Cloudless Day",
  booktitle="European Conference on Computer Vision",
  year="2010",
  pages="368--381",
}

@article{agarwal2011building,
  author = {Agarwal, Sameer and Furukawa, Yasutaka and Snavely, Noah and Simon, Ian and Curless, Brian and Seitz, Steven M. and Szeliski, Richard},
  title = {Building Rome in a day},
  year = {2011},
  volume = {54},
  number = {10},
  journal = {Communications of the ACM},
  month = oct,
  pages = {105–112},
}

@INPROCEEDINGS{wu2013towards,
  author={Wu, Changchang},
  booktitle={2013 International Conference on 3D Vision}, 
  title={Towards Linear-Time Incremental Structure from Motion}, 
  year={2013},
  volume={},
  number={},
  pages={127-134},
  }

@InProceedings{schoenberger2016sfm,
  author = {Schonberger, Johannes L. and Frahm, Jan-Michael},
  title = {{Structure-From-Motion Revisited}},
  booktitle = {Proceedings of the IEEE Conference on Computer Vision and Pattern Recognition},
  month = {June},
  year = {2016}
}

@ARTICLE{crandall2012sfm,
  author={Crandall, David J. and Owens, Andrew and Snavely, Noah and Huttenlocher, Daniel P.},
  journal={IEEE Transactions on Pattern Analysis and Machine Intelligence}, 
  title={SfM with MRFs: Discrete-Continuous Optimization for Large-Scale Structure from Motion}, 
  year={2013},
  volume={35},
  number={12},
  pages={2841-2853},
}

@InProceedings{yi_lift_2016,
  author="Yi, Kwang Moo
  and Trulls, Eduard
  and Lepetit, Vincent
  and Fua, Pascal",
  title="LIFT: Learned Invariant Feature Transform",
  booktitle="European Conference on Computer Vision",
  year="2016",
  pages="467--483",
}

@InProceedings{detone2018self,
author = {DeTone, Daniel and Malisiewicz, Tomasz and Rabinovich, Andrew},
title = {SuperPoint: Self-Supervised Interest Point Detection and Description},
booktitle = {Proceedings of the IEEE Conference on Computer Vision and Pattern Recognition Workshops},
month = {June},
year = {2018}
}

@InProceedings{dusmanu2019d2,
  author = {Dusmanu, Mihai and Rocco, Ignacio and Pajdla, Tomas and Pollefeys, Marc and Sivic, Josef and Torii, Akihiko and Sattler, Torsten},
  title = {{D2-Net: A Trainable CNN for Joint Detection and Description of Local Features}},
  booktitle = {Proceedings of the IEEE/CVF Conference on Computer Vision and Pattern Recognition},
  year = {2019},
}

@inproceedings{tyszkiewicz2020disk,
  author = {Tyszkiewicz, Micha\l  and Fua, Pascal and Trulls, Eduard},
  booktitle = {Advances in Neural Information Processing Systems},
  pages = {14254--14265},
  title = {{DISK: Learning local features with policy gradient}},
  volume = {33},
  year = {2020}
}

@INPROCEEDINGS{sarlin2020superglue,
  author={Sarlin, Paul-Edouard and DeTone, Daniel and Malisiewicz, Tomasz and Rabinovich, Andrew},
  booktitle={Proceedings of the IEEE/CVF Conference on Computer Vision and Pattern Recognition}, 
  title={{SuperGlue: Learning Feature Matching With Graph Neural Networks}}, 
  year={2020},
  volume={},
  number={},
  pages={4937-4946},
  }

@InProceedings{sun2021loftr,
    author    = {Sun, Jiaming and Shen, Zehong and Wang, Yuang and Bao, Hujun and Zhou, Xiaowei},
    title     = {{LoFTR: Detector-Free Local Feature Matching With Transformers}},
    booktitle = {Proceedings of the IEEE/CVF Conference on Computer Vision and Pattern Recognition},
    month     = {June},
    year      = {2021},
    pages     = {8922-8931}
}

@InProceedings{chen2022aspanformer,
  author="Chen, Hongkai
  and Luo, Zixin
  and Zhou, Lei
  and Tian, Yurun
  and Zhen, Mingmin
  and Fang, Tian
  and McKinnon, David
  and Tsin, Yanghai
  and Quan, Long",
  title="ASpanFormer: Detector-Free Image Matching with Adaptive Span Transformer",
  booktitle="European Conference on Computer Vision",
  year="2022",
  pages="20--36",
}

@InProceedings{lindenberger2023lightglue,
    author    = {Lindenberger, Philipp and Sarlin, Paul-Edouard and Pollefeys, Marc},
    title     = {{LightGlue: Local Feature Matching at Light Speed}},
    booktitle = {Proceedings of the IEEE/CVF International Conference on Computer Vision},
    month     = {October},
    year      = {2023},
    pages     = {17627-17638}
}

@ARTICLE{truong2023pdc,
  author={Truong, Prune and Danelljan, Martin and Timofte, Radu and Van Gool, Luc},
  journal={IEEE Transactions on Pattern Analysis and Machine Intelligence}, 
  title={{PDC-Net+: Enhanced Probabilistic Dense Correspondence Network}}, 
  year={2023},
  volume={45},
  number={8},
  pages={10247-10266},
}

@ARTICLE{hartley1997defense,
  author={Hartley, R.I.},
  journal={IEEE Transactions on Pattern Analysis and Machine Intelligence}, 
  title={In defense of the eight-point algorithm}, 
  year={1997},
  volume={19},
  number={6},
  pages={580-593},
}

@ARTICLE{quan1999linear,
  author={Long Quan and Zhongdan Lan},
  journal={IEEE Transactions on Pattern Analysis and Machine Intelligence}, 
  title={{Linear N-point camera pose determination}}, 
  year={1999},
  volume={21},
  number={8},
  pages={774-780},
  }

@ARTICLE{gao2003complete,
  author={Xiao-Shan Gao and Xiao-Rong Hou and Jianliang Tang and Hang-Fei Cheng},
  journal={IEEE Transactions on Pattern Analysis and Machine Intelligence}, 
  title={Complete solution classification for the perspective-three-point problem}, 
  year={2003},
  volume={25},
  number={8},
  pages={930-943},
  }

@ARTICLE{nister2004efficient,
  author={Nister, D.},
  journal={IEEE Transactions on Pattern Analysis and Machine Intelligence}, 
  title={{An efficient solution to the five-point relative pose problem}}, 
  year={2004},
  volume={26},
  number={6},
  pages={756-770},
  }

@ARTICLE{umeyama1991least,
  author={Umeyama, S.},
  journal={IEEE Transactions on Pattern Analysis and Machine Intelligence}, 
  title={{Least-squares estimation of transformation parameters between two point patterns}}, 
  year={1991},
  volume={13},
  number={4},
  pages={376-380},
  }

@ARTICLE{besl1992icp,
  author={Besl, P.J. and McKay, Neil D.},
  journal={IEEE Transactions on Pattern Analysis and Machine Intelligence}, 
  title={A method for registration of 3-D shapes}, 
  year={1992},
  volume={14},
  number={2},
  pages={239-256},
  }

@ARTICLE{Furukawa2010MVS,
  author={Furukawa, Yasutaka and Ponce, Jean},
  journal={IEEE Transactions on Pattern Analysis and Machine Intelligence}, 
  title={Accurate, Dense, and Robust Multiview Stereopsis}, 
  year={2010},
  volume={32},
  number={8},
  pages={1362-1376},
  }

@InProceedings{galliani2015massively,
author = {Galliani, Silvano and Lasinger, Katrin and Schindler, Konrad},
title = {Massively Parallel Multiview Stereopsis by Surface Normal Diffusion},
booktitle = {Proceedings of the IEEE International Conference on Computer Vision},
month = {December},
year = {2015}
}

@inproceedings{schonberger2016pixelwise,
  author={Johannes L. Schönberger and Enliang Zheng and Jan-Michael Frahm and Marc Pollefeys},
  title={{Pixelwise View Selection for Unstructured Multi-View Stereo}},
  year={2016},
  pages={501-518},
  booktitle={European Conference on Computer Vision}
}

@InProceedings{yao2018mvsnet,
author = {Yao, Yao and Luo, Zixin and Li, Shiwei and Fang, Tian and Quan, Long},
title = {{MVSNet: Depth Inference for Unstructured Multi-view Stereo}},
booktitle = {European Conference on Computer Vision},
month = {September},
year = {2018}
}

@InProceedings{gu2020cascade,
  author = {Gu, Xiaodong and Fan, Zhiwen and Zhu, Siyu and Dai, Zuozhuo and Tan, Feitong and Tan, Ping},
  title = {{Cascade Cost Volume for High-Resolution Multi-View Stereo and Stereo Matching}},
  booktitle = {Proceedings of the IEEE/CVF Conference on Computer Vision and Pattern Recognition},
  month = {June},
  year = {2020}
}

@InProceedings{geomvsnet,
    author    = {Zhang, Zhe and Peng, Rui and Hu, Yuxi and Wang, Ronggang},
    title     = {{GeoMVSNet: Learning Multi-View Stereo With Geometry Perception}},
    booktitle = {Proceedings of the IEEE/CVF Conference on Computer Vision and Pattern Recognition},
    month     = {June},
    year      = {2023},
    pages     = {21508-21518}
}

@article{xu2022multi,
  title={{Multi-Scale Geometric Consistency Guided and Planar Prior Assisted Multi-View Stereo}}, 
  author={Xu, Qingshan and Kong, Weihang and Tao, Wenbing and Pollefeys, Marc}, 
  journal={IEEE Transactions on Pattern Analysis and Machine Intelligence}, 
  volume={45}, 
  number={4}, 
  pages={4945--4963}, 
  year={2022}
}

@article{mildenhall2021nerf,
author = {Mildenhall, Ben and Srinivasan, Pratul P. and Tancik, Matthew and Barron, Jonathan T. and Ramamoorthi, Ravi and Ng, Ren},
title = {{NeRF: representing scenes as neural radiance fields for view synthesis}},
year = {2021},
volume = {65},
number = {1},
journal = {Communications of the ACM},
month = dec,
pages = {99–106},
}

@article{kerbl3Dgaussians,
  title={{3D Gaussian Splatting for Real-Time Radiance Field Rendering}}, 
  author={Kerbl, Bernhard and Kopanas, Georgios and Leimk{"u}hler, George and Drettakis, George}, 
  journal={ACM Transactions on Graphics}, volume={42}, number={4}, year={2023}
}

@InProceedings{zhou2017unsupervised,
  author = {Zhou, Tinghui and Brown, Matthew and Snavely, Noah and Lowe, David G.},
  title = {{Unsupervised Learning of Depth and Ego-Motion From Video}},
  booktitle = {Proceedings of the IEEE Conference on Computer Vision and Pattern Recognition},
  month = {July},
  year = {2017}
}

@InProceedings{ummenhofer2017demon,
author = {Ummenhofer, Benjamin and Zhou, Huizhong and Uhrig, Jonas and Mayer, Nikolaus and Ilg, Eddy and Dosovitskiy, Alexey and Brox, Thomas},
title = {{DeMoN: Depth and Motion Network for Learning Monocular Stereo}},
booktitle = {Proceedings of the IEEE Conference on Computer Vision and Pattern Recognition},
month = {July},
year = {2017}
}

@inproceedings{tang2018ba,
  title={{BA}-Net: Dense Bundle Adjustment Networks},
  author={Chengzhou Tang and Ping Tan},
  booktitle={International Conference on Learning Representations},
  year={2019},
}

@inproceedings{wei2020deepsfm,
  title={{DeepSFM: Structure From Motion via Deep Bundle Adjustment}}, author={Wei, Xingkui and Zhang, Yinda and Li, Zhuwen}, 
  booktitle={European Conference on Computer Vision}, year={2020}
}

@inproceedings{teed2018deepv2d,
  title={{DeepV2D: Video to Depth with Differentiable Structure from Motion}},
  author={Zachary Teed and Jia Deng},
  booktitle={International Conference on Learning Representations},
  year={2020},
}

@inproceedings{teed2021droid,
 author = {Teed, Zachary and Deng, Jia},
 booktitle = {Advances in Neural Information Processing Systems},
 pages = {16558--16569},
 title = {DROID-SLAM: Deep Visual SLAM for Monocular, Stereo, and \mbox{RGB-D} Cameras},
 volume = {34},
 year = {2021}
}

@InProceedings{dai2017scannet,
author = {Dai, Angela and Chang, Angel X. and Savva, Manolis and Halber, Maciej and Funkhouser, Thomas and Niessner, Matthias},
title = {{ScanNet: Richly-Annotated 3D Reconstructions of Indoor Scenes}},
booktitle = {Proceedings of the IEEE Conference on Computer Vision and Pattern Recognition},
month = {July},
year = {2017}
}

@InProceedings{shotton2013scene,
author = {Shotton, Jamie and Glocker, Ben and Zach, Christopher and Izadi, Shahram and Criminisi, Antonio and Fitzgibbon, Andrew},
title = {{Scene Coordinate Regression Forests for Camera Relocalization in RGB-D Images}},
booktitle = {Proceedings of the IEEE Conference on Computer Vision and Pattern Recognition},
month = {June},
year = {2013}
}

@InProceedings{Butler:ECCV:2012,
author="Butler, Daniel J.
and Wulff, Jonas
and Stanley, Garrett B.
and Black, Michael J.",
title="A Naturalistic Open Source Movie for Optical Flow Evaluation",
booktitle="European Conference on Computer Vision",
year="2012",
pages="611--625",
}

@INPROCEEDINGS{palazzolo2019iros,
  author={Palazzolo, Emanuele and Behley, Jens and Lottes, Philipp and Giguère, Philippe and Stachniss, Cyrill},
  booktitle={IEEE/RSJ International Conference on Intelligent Robots and Systems}, 
  title={{ReFusion: 3D Reconstruction in Dynamic Environments for RGB-D Cameras Exploiting Residuals}}, 
  year={2019},
  volume={},
  number={},
  pages={7855-7862},
  }

@inproceedings{gao2022dynamic,
 author = {Gao, Hang and Li, Ruilong and Tulsiani, Shubham and Russell, Bryan and Kanazawa, Angjoo},
 booktitle = {Advances in Neural Information Processing Systems},
 pages = {33768--33780},
 title = {Monocular Dynamic View Synthesis: A Reality Check},
 volume = {35},
 year = {2022}
}

@InProceedings{Azinovic_2022_CVPR,
    author    = {Azinovi\'c, Dejan and Martin-Brualla, Ricardo and Goldman, Dan B and Nie{\ss}ner, Matthias and Thies, Justus},
    title     = {{Neural RGB-D Surface Reconstruction}},
    booktitle = {Proceedings of the IEEE/CVF Conference on Computer Vision and Pattern Recognition},
    month     = {June},
    year      = {2022},
    pages     = {6290-6301}
}

@ARTICLE{mikolajczyk2005evaluation,
  author={Mikolajczyk, K. and Schmid, C.},
  journal={IEEE Transactions on Pattern Analysis and Machine Intelligence}, 
  title={A performance evaluation of local descriptors}, 
  year={2005},
  volume={27},
  number={10},
  pages={1615-1630},
  }

@ARTICLE{liu2010sift,
  author={Liu, Ce and Yuen, Jenny and Torralba, Antonio},
  journal={IEEE Transactions on Pattern Analysis and Machine Intelligence}, 
  title={{SIFT Flow: Dense Correspondence across Scenes and Its Applications}}, 
  year={2011},
  volume={33},
  number={5},
  pages={978-994},
  }

@ARTICLE{dso,
  author={Engel, Jakob and Koltun, Vladlen and Cremers, Daniel},
  journal={IEEE Transactions on Pattern Analysis and Machine Intelligence}, 
  title={Direct Sparse Odometry}, 
  year={2018},
  volume={40},
  number={3},
  pages={611-625},
  }

@ARTICLE{torresani2008nonrigid,
  author={Torresani, Lorenzo and Hertzmann, Aaron and Bregler, Chris},
  journal={IEEE Transactions on Pattern Analysis and Machine Intelligence}, 
  title={{Nonrigid Structure-from-Motion: Estimating Shape and Motion with Hierarchical Priors}}, 
  year={2008},
  volume={30},
  number={5},
  pages={878-892},
  }

@ARTICLE{liao2022kitti,
  author={Liao, Yiyi and Xie, Jun and Geiger, Andreas},
  journal={IEEE Transactions on Pattern Analysis and Machine Intelligence}, 
  title={{KITTI-360: A Novel Dataset and Benchmarks for Urban Scene Understanding in 2D and 3D}}, 
  year={2023},
  volume={45},
  number={3},
  pages={3292-3310},
  }

@ARTICLE{fan2019localfeatures,
  author={Fan, Bin and Kong, Qingqun and Wang, Xinchao and Wang, Zhiheng and Xiang, Shiming and Pan, Chunhong and Fua, Pascal},
  journal={IEEE Transactions on Image Processing}, 
  title={{A Performance Evaluation of Local Features for Image-Based 3D Reconstruction}}, 
  year={2019},
  volume={28},
  number={10},
  pages={4774-4789},
  }

@ARTICLE{shen2023semidense,
  author={Shen, Zehong and Sun, Jiaming and Wang, Yuang and He, Xingyi and Bao, Hujun and Zhou, Xiaowei},
  journal={IEEE Transactions on Pattern Analysis and Machine Intelligence}, 
  title={{Semi-Dense Feature Matching With Transformers and its Applications in Multiple-View Geometry}}, 
  year={2023},
  volume={45},
  number={6},
  pages={7726-7738},
  }

@ARTICLE{sarlin2025pixsfm,
  author={Sarlin, Paul-Edouard and Lindenberger, Philipp and Larsson, Viktor and Pollefeys, Marc},
  journal={IEEE Transactions on Pattern Analysis and Machine Intelligence}, 
  title={{Pixel-Perfect Structure-From-Motion With Featuremetric Refinement}}, 
  year={2025},
  volume={47},
  number={5},
  pages={3298-3309},
  }

@ARTICLE{shen2013patchbased,
  author={Shen, Shuhan},
  journal={IEEE Transactions on Image Processing}, 
  title={{Accurate Multiple View 3D Reconstruction Using Patch-Based Stereo for Large-Scale Scenes}}, 
  year={2013},
  volume={22},
  number={5},
  pages={1901-1914},
  }

@ARTICLE{li2020cldmvs,
  author={Li, Zhaoxin and Zuo, Wangmeng and Wang, Zhaoqi and Zhang, Lei},
  journal={IEEE Transactions on Image Processing}, 
  title={{Confidence-Based Large-Scale Dense Multi-View Stereo}}, 
  year={2020},
  volume={29},
  number={},
  pages={7176-7191},
  }

@ARTICLE{li2023nrmvsnet,
  author={Li, Jingliang and Lu, Zhengda and Wang, Yiqun and Xiao, Jun and Wang, Ying},
  journal={IEEE Transactions on Image Processing}, 
  title={{NR-MVSNet: Learning Multi-View Stereo Based on Normal Consistency and Depth Refinement}}, 
  year={2023},
  volume={32},
  number={},
  pages={2649-2662},
  }

@ARTICLE{ji2021surfacenet,
  author={Ji, Mengqi and Zhang, Jinzhi and Dai, Qionghai and Fang, Lu},
  journal={IEEE Transactions on Pattern Analysis and Machine Intelligence}, 
  title={{SurfaceNet+: An End-to-end 3D Neural Network for Very Sparse Multi-View Stereopsis}}, 
  year={2021},
  volume={43},
  number={11},
  pages={4078-4093},
  }

@ARTICLE{zhang2022gigamvs,
  author={Zhang, Jianing and Zhang, Jinzhi and Mao, Shi and Ji, Mengqi and Wang, Guangyu and Chen, Zequn and Zhang, Tian and Yuan, Xiaoyun and Dai, Qionghai and Fang, Lu},
  journal={IEEE Transactions on Pattern Analysis and Machine Intelligence}, 
  title={{GigaMVS: A Benchmark for Ultra-Large-Scale Gigapixel-Level 3D Reconstruction}}, 
  year={2022},
  volume={44},
  number={11},
  pages={7534-7550},
  }

@article{oquab2024dinov,
  title={{DINO}v2: Learning Robust Visual Features without Supervision},
  author={Maxime Oquab and Timoth{\'e}e Darcet and Th{\'e}o Moutakanni and Huy V. Vo and Marc Szafraniec and Vasil Khalidov and Pierre Fernandez and Daniel HAZIZA and Francisco Massa and Alaaeldin El-Nouby and Mido Assran and Nicolas Ballas and Wojciech Galuba and Russell Howes and Po-Yao Huang and Shang-Wen Li and Ishan Misra and Michael Rabbat and Vasu Sharma and Gabriel Synnaeve and Hu Xu and Herve Jegou and Julien Mairal and Patrick Labatut and Armand Joulin and Piotr Bojanowski},
  journal={Transactions on Machine Learning Research},
  year={2024},
}

@inproceedings{ICLR2026_3d551707,
 author = {Zhou, Kaichen and Wang, Yuhan and Chen, Grace and Beaudouin, Gaspard and Zhan, Fangneng and Liang, Paul and Wang, Mengyu},
 booktitle = {International Conference on Learning Representations},
 pages = {36401--36414},
 title = {PAGE-4D: Disentangled Pose and Geometry Estimation for VGGT-4D Perception},
 year = {2026}
}




%








\end{document}